\pdfoutput=1

\documentclass[11pt]{article}
\usepackage{algorithm}
\usepackage{algorithmicx}
\usepackage{algorithm,algpseudocode}
\usepackage{amsmath}
\usepackage[preprint]{acl}

\usepackage{times}
\usepackage{latexsym}
\usepackage{multirow}
\usepackage{colortbl}
\usepackage{subcaption}
\usepackage{booktabs}
\usepackage[T1]{fontenc}

\usepackage[utf8]{inputenc}

\usepackage{microtype}

\usepackage{inconsolata}

\usepackage{graphicx}
\usepackage{subcaption}
\usepackage{caption}
\usepackage{makecell}

\title{When Facts Change: Probing LLMs on Evolving Knowledge with \texttt{evolveQA}}

\author{
Nishanth Sridhar Nakshatri\thanks{Work done during an internship at AWS AI Labs.}\textsuperscript{$\clubsuit$} \quad
Shamik Roy \textsuperscript{$\spadesuit$} \quad
Manoj Ghuhan Arivazhagan\textsuperscript{$\spadesuit$} \quad \\
\textbf{Hanhan Zhou}\textsuperscript{$\spadesuit$} \quad
    \textbf{Vinayshekhar Bannihatti Kumar}\textsuperscript{$\spadesuit$}\quad
    \textbf{Rashmi Gangadharaiah}\textsuperscript{$\spadesuit$}\quad
    \vspace{0.1in} \\
    \textsuperscript{$\clubsuit$}Purdue University 
    \textsuperscript{$\spadesuit$}AWS AI Labs\\
    {\tt nnakshat@purdue.edu}\\
    {\tt \{royshami, mghuhan, hanhanz, vinayshk, rgangad\}@amazon.com}
}

\begin{document}
\maketitle
\begin{abstract}
LLMs often fail to handle temporal knowledge conflicts--contradictions arising when facts evolve over time within their training data. Existing studies evaluate this phenomenon through benchmarks built on structured knowledge bases like Wikidata, but they focus on widely-covered, easily-memorized popular entities and lack the dynamic structure needed to fairly evaluate LLMs with different knowledge cut-off dates. We introduce \texttt{evolveQA}%
\footnote{Code and data coming soon.}, a benchmark specifically designed to evaluate LLMs on temporally evolving knowledge, constructed from $3$ real-world, time-stamped corpora: AWS updates, Azure changes, and WHO disease outbreak reports. Our framework identifies naturally occurring knowledge evolution and generates questions with gold answers tailored to different LLM knowledge cut-off dates. Through extensive evaluation of $12$ open and closed-source LLMs across $3$ knowledge probing formats, we demonstrate significant performance drops of up to $31$\% on \texttt{evolveQA} compared to static knowledge questions.

\end{abstract}

\vspace{-0.4cm}
\section{Introduction}
\begin{figure}[t!]
\includegraphics[width=\linewidth]{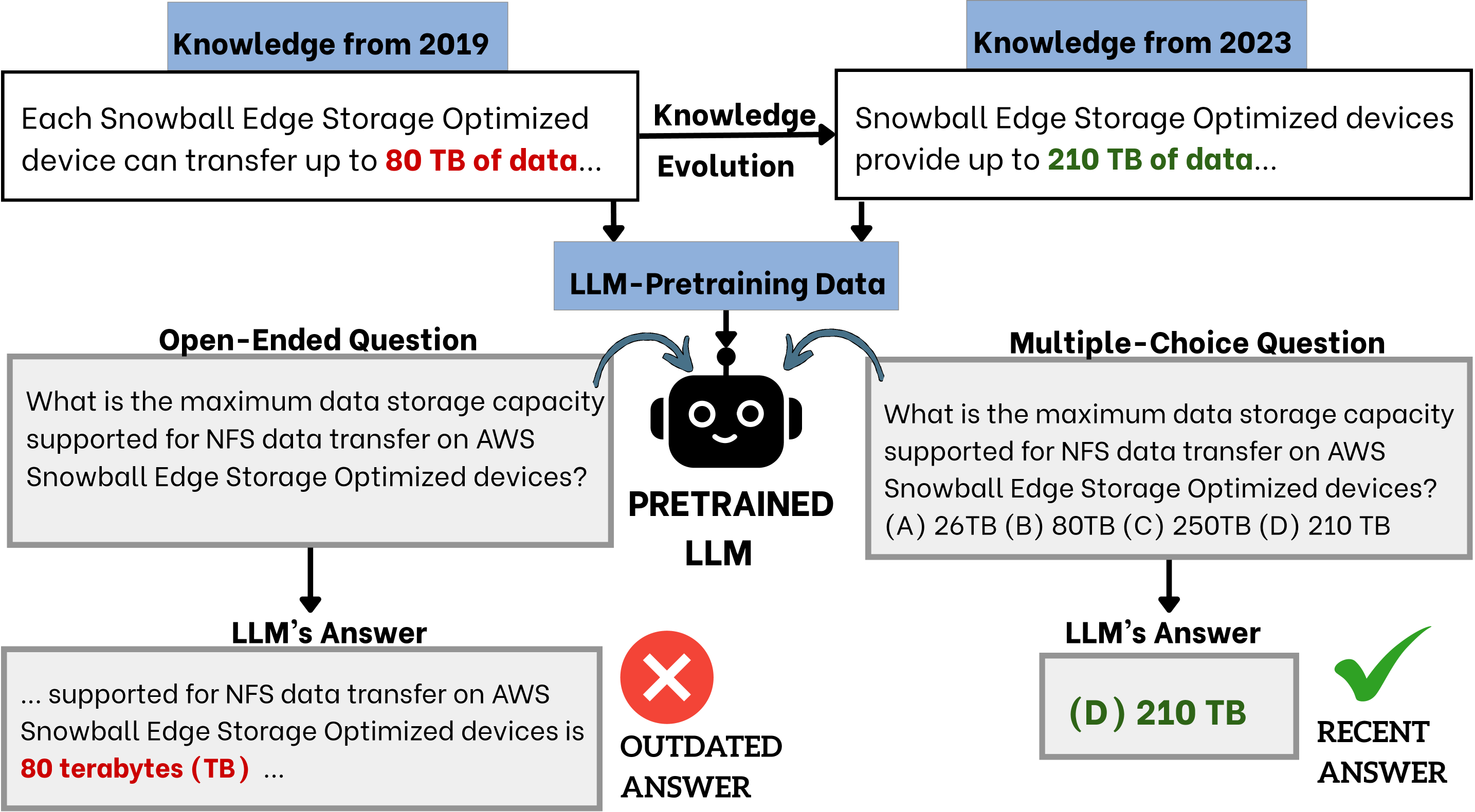}
\caption{\small LLMs often give outdated answers to open-ended questions, even though they can identify correct and current information when given multiple-choice options; this suggests that while LLMs hold more recent knowledge in their parameters, they struggle to recall it without explicit prompting.}
\label{fig:intro-image}
\vspace{-0.5cm}
\end{figure}

Large Language Models (LLMs) have demonstrated remarkable capabilities across diverse tasks, from text generation to complex reasoning~\cite{brown2020language, wei2021finetuned, chang2024survey, su2024living}. However, their reliability faces a critical challenge: \textit{temporal knowledge conflicts}, where factual information evolves over time, creating contradictions between outdated and current versions within the model's training data. This problem is pervasive across domains such as healthcare, where guidelines update with new research; technology, where specifications change with software releases, and so on. As LLMs increasingly serve as knowledge sources in high-stake applications, their tendency to recall outdated information over recent updates, as illustrated in Figure~\ref{fig:intro-image}, poses significant risks to system reliability and user trust.

Prior works studied broadly three types of knowledge conflicts in LLMs: conflicts in external retrieved documents~\cite{pan2021attacking, chen2022rich, wang2025astuteragovercomingimperfect}, conflicts between external context and model's internal knowledge~\cite{kortukov2024studying, wang2023resolving, xie2023adaptive, neeman2022disentqa}, and conflicts within an LLM's parametric knowledge~\cite{elazar-etal-2021-measuring, elsahar-etal-2018-rex, dong2023statistical, zhao2023knowing}, often stemming from noise, biases, or outdated information in the vast training corpora. %

However, existing approaches face critical limitations when addressing temporal knowledge conflicts, a severely under-explored area of parametric knowledge conflicts. Current benchmarks rely heavily on popular entities from structured knowledge bases like Wikidata~\cite{kim2023carpe, marjanovic2024dynamicqa, su2024conflictbank, ozer2025question, tang-etal-2025-evowiki}, leading to inadequate evaluation due to LLMs' strong memorization of such entities~\cite{mallen2022not}. Moreover, the static nature of most benchmarks hinders a fair evaluation of LLMs with differing knowledge cut-off dates~\cite{tang-etal-2025-evowiki}. Additionally, ~\cite{ozer2025question} employ Subject-Relation-Object (SRO) triplet-based extraction methods that fail to capture nuanced contextual changes in complex corpora, and often resort to artificially induced conflicts~\cite{su2024conflictbank} rather than capturing how temporal conflicts naturally arise in real-world text. These limitations are particularly problematic because temporal conflicts represent one of the most prevalent forms of knowledge inconsistency in practice~\cite{kaddour2023challenges, zhou2023lima}, yet our understanding of how LLMs handle naturally occurring temporal evolution remains severely limited.

To address these limitations, we propose a novel framework for a systematic benchmark construction, capturing temporally evolving knowledge from unstructured, time-stamped corpora (Section \ref{sec:framework}). Our methodology identifies naturally occurring knowledge evolution through a multi-stage process: (1) extracting salient entities and concepts from documents, (2) clustering concepts to identify overarching topics, (3) isolating specific attributes that change over time within entity-concept pairs, and (4) generating grounded questions with time-sensitive gold answers tailored to different LLM knowledge cut-off dates. Unlike previous studies that rely on synthetic conflicts or structured knowledge bases, our approach leverages real-world data sources where information genuinely evolves, providing authentic evaluation of temporal knowledge handling capabilities.

Using our framework, we introduce \texttt{evolveQA}, a comprehensive question answering (QA) benchmark for evolving knowledge. It is constructed across $3$ dynamic domains: AWS service updates ($550$ QA pairs), Azure platform changes ($203$ QA pairs), and WHO-Disease Outbreak News reports ($169$ QA pairs), spanning multiple temporal cut-off dates and $3$ question formats: open-ended, multiple-choice, and verifiable QA (Section \ref{sec:evolveQA}). Our extensive evaluation of $12$ LLMs ($6$ open-source, $6$ closed-source) from $4$ model families reveals universal and significant performance degradation on temporally evolving knowledge, with accuracy drops of $6\% - 31\%$ across domains compared to non-conflicting scenarios. Critically, we demonstrate that question format substantially impacts performance: while models achieve $53\% - 76\%$ accuracy on MCQ questions about evolving facts, they drop to $12\% - 51\%$ on open-ended queries targeting the same knowledge. Interestingly, in $32\%-45\%$ of cases, models generate outdated responses to open-ended questions but select the correct option in the corresponding MCQ format, indicating that updated knowledge exists in parametric memory but is poorly prioritized during recall (Section \ref{sec:experimental-eval}). Such findings provide crucial insights into the nature of temporal knowledge conflicts and establish a foundation for developing more temporally robust LLMs.

\section{Temporally Evolving Knowledge Benchmark Creation Framework}
\label{sec:framework}

\begin{figure*}[t!]
\centering
\includegraphics[height=8.3cm, width=\textwidth]{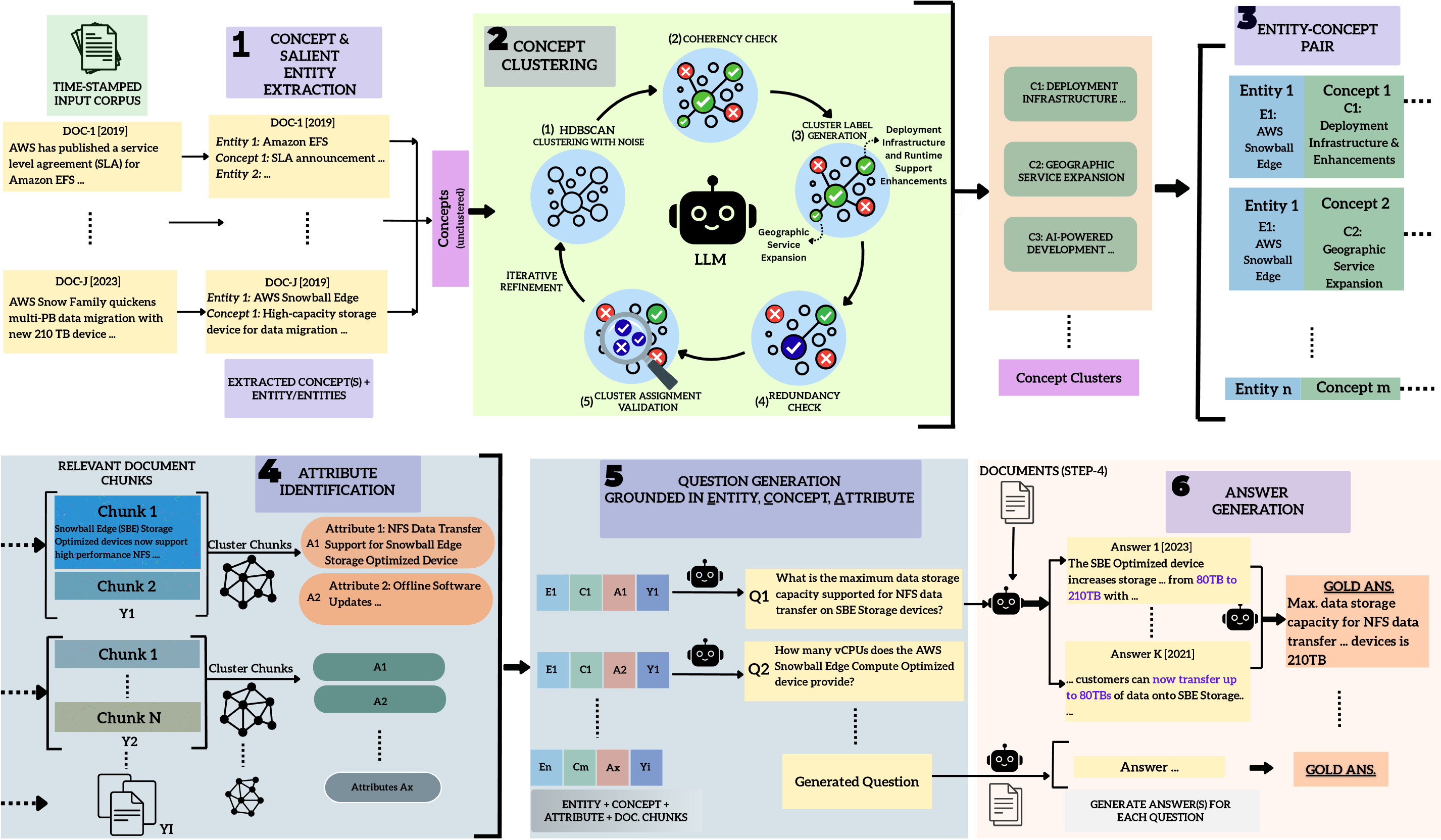}

\caption{\small Framework for \texttt{evolveQA} construction. It operates in two phases: \textbf{1. Evolving Fact Identification:} (1) Extract entities and concepts, (2) cluster concepts into topics, (3) obtain documents for each \texttt{\{Entity, Concept\}} pair, (4) cluster document chunks to identify temporal attributes. \textbf{2. Grounded QA Curation:} (5) Generate specific questions and (6) curate time-sensitive gold answers for each \texttt{\{Entity, Concept, Attribute\}} tuple.}

  \vspace{-0.4cm}
  \label{fig:main-framework-image}
\end{figure*}

In this section, we propose a novel framework to identify temporally evolving knowledge from a given text corpora and develop a benchmark of question-answer pairs to probe LLMs' understanding of this knowledge. This framework involves two major steps: (1) systematically navigating a corpus of time-stamped documents to \textit{identify temporally evolving facts}; (2) \textit{generating high-quality questions} that query LLMs for these evolving facts, along with their corresponding \textit{gold-standard answers} tailored to different LLM knowledge cut-off dates. Our approach is outlined in Figure~\ref{fig:main-framework-image}.

\subsection{Identifying Temporally Evolving Facts}
Our goal is to first identify evolving facts from a given text corpus where the most recent information is considered authoritative. To achieve this, we employ an LLM-based approach as follows.

\subsubsection{Concept and Salient Entity Extraction}
We begin with a text corpus tagged with publication timestamps. To identify the precise knowledge that evolves over time in this corpus, we first need to group documents discussing similar topics. However, real-world documents are inherently noisy, and often cover multiple sub-topics, making it difficult to cluster the documents into meaningful groups. To mitigate this, following previous works~\cite{nakshatri2025talking, tamkin2024clio}, we rely on clustering abstract representations of the documents instead of the raw documents. In particular, we first distill documents into their key semantic components, which act as discriminative anchors that enable effective grouping of related documents. We define such a component as the combination of the following$-$ (a) \textit{concept:} a short text description that provides an abstract, yet informative view of a central discussion point in the document. (b) \textit{entity}: salient subject associated with the identified concept.

For every document in the given time-stamped corpus, we use an LLM to extract up to $k$ %
high-level \textit{concepts} along with their corresponding \textit{salient entities}. This pairing of an entity with its associated context provides a robust, topic-centric representation of the document's content~\cite{chen2023towards, nanni2017building}, forming a solid foundation for subsequent clustering. The \textit{step-1} in Figure~\ref{fig:main-framework-image} shows an example of a salient entity: ``AWS Snowball Edge'', and a concept: ``High-capacity storage device for data migration'', extracted from a document in the AWS updates corpus.

\vspace{-0.1cm}
\subsubsection{High-Coherence Concept Clustering} 
In this step, we cluster the \textit{concepts} extracted in the previous step rather than the raw documents. This approach enables embedding-based clustering algorithms to effectively identify semantically dense regions in the embedding space, leading to coherent initial topic groups. These topics are further refined iteratively with an LLM-in-the-loop approach. Unlike prior works that primarily use LLMs for cluster labeling and refinement~\cite{nakshatri2023using}, or preference alignment~\cite{zhang2023clusterllm}, our method leverages LLMs both (a) \textit{during} the clustering process to label and refine the generated concept clusters, and (b) \textit{post clustering} to validate and correct concept assignments. Our iterative process works as follows.

\textbf{Step 1: Concept-Label Discovery.} We begin by embedding all extracted concepts using a text embedding model and then apply HDBSCAN algorithm~\cite{campello2013density} to identify dense regions in the embedding space. Then we perform three critical LLM-guided refinement steps.

\begin{itemize}
    \vspace{-0.3cm}
    \item \textbf{Coherency check:} To ensure high cluster coherence, we use an LLM to validate whether the top-$m$ ($m=5$) most central members within each cluster are semantically aligned with each other. If these members do not consistently discuss the same overall concept, the cluster is marked as \textit{incoherent}, hence, dissolved and its constituent concepts are returned to an unclustered pool. %
    \vspace{-0.3cm}
    \item \textbf{Cluster-label generation:} For each coherent cluster, an LLM is prompted to assign a concise label based on the semantic content of its top-$m$ documents. %
    This label serves as the overarching concept discussed in the cluster. 
    \vspace{-0.3cm}
    \item \textbf{Redundancy check:} %
    To avoid multiple clusters discussing the same concept, an LLM is used to validate if a cluster pair with semantically similar cluster labels (determined based on cosine similarity between cluster label embeddings) can be merged, ensuring conceptually distinct final clusters. %
\end{itemize}
\vspace{-0.3cm}

\textbf{Step 2: Concept-Cluster Assignment Validation.} In this step, we validate individual concept assignments. First, within each cluster, an LLM verifies if every concept is consistent with the assigned cluster label. Concepts deemed inconsistent are removed and returned to the unclustered pool. Subsequently, for any remaining unclustered concepts, an LLM attempts to assign them to one of the top-$3$ existing, cluster labels (obtained based on cosine similarity between the concept and cluster label embeddings). This process ensures the alignment of each concept in a cluster with its label. %

\textbf{Step 3: Iterative Refinement.}  If a significant number of concepts remain unclustered, we return to \textbf{Step 1} and repeat the process. %
This iterative loop continues until all concepts are clustered or a maximum iteration count (set to $5$) is reached. This iterative, LLM-guided process ensures that the final concept clusters are both highly coherent and conceptually distinct, laying a solid foundation for identifying evolving knowledge attributes. %

\subsubsection{Identifying Evolving Attributes}
The generated concept cluster labels from the previous step are abstract in nature. For instance, ``AI-Powered Development and Integration, Geographic Service Expansion etc.'' (shown as the output of \textit{step-2} in Figure~\ref{fig:main-framework-image}), are too abstract for generating precise questions about changing facts. One cannot ask a specific, answerable question based on such a broad theme alone. Therefore, in this step, by organizing the source corpora based on a \{\textit{entity, concept cluster label}\} pair, we pinpoint the specific, fine-grained \textit{attributes} within these documents that genuinely evolve over time. An \textit{attribute} represents a specific property or detail, such as ``Offline Software Updates for Air-Gapped Snowball Edge Devices'', that is susceptible to temporal change.

To achieve this, we utilize the structure developed in the previous steps. For a concept cluster, we begin by aggregating all the (\textit{entity, document}) pairs that correspond to the concepts within that cluster. This process naturally groups documents into topically focused subsets of the corpus, i.e, all documents related to an entity, say ``AWS Snowball Edge'', and a concept, say ``Deployment Infrastructure and Enhancements'', are grouped together.

Within this document subset, our goal is to isolate the discrete factual statements that undergo updates. Since documents may contain multiple attributes, we first segment them into smaller units (or chunks). We employ a sliding window of three sentences per chunk with one-sentence overlap to maintain contextual integrity. These chunks are then embedded and clustered using HBDSCAN algorithm. Each resulting cluster of chunks now represents a recurring, specific piece of information. An LLM is prompted to assign a descriptive label to each chunk cluster, which is referred to as the \textit{attribute}. Formally, all document chunks within a single \{\textit{entity, concept cluster label, attribute}\} tuple represent the same specific piece of potentially evolving knowledge. 

This three-part tuple forms the fundamental unit for our question generation. For brevity in the subsequent sections, we will refer to the \{\textit{concept cluster label}\} simply as the \{\textit{concept}\}.

\vspace{-0.2cm}
\subsection{Question \& Answer Generation}
After identification of evolving knowledge \textit{attributes}, here, we generate questions that target them and curate time-sensitive gold-standard answers for various LLM knowledge cut-off dates.

\subsubsection{Question Generation} 
For question generation, we consider a specific cluster corresponding to a unique \texttt{(entity , concept, attribute)} tuple. To ensure questions are well-grounded, we sample the top-$K$ chunks from different documents and from varying timestamps within this \textit{attribute} cluster (based on cosine similarity to the cluster centroid). Using these selected chunks as contextual anchors, we prompt an LLM to formulate a question that is explicitly grounded in the three dimensions of \texttt{(entity , concept, attribute)} tuple. This grounding is crucial for ensuring precision and query disambiguation. The \textit{entity}, say ``AWS Snowball Edge'', establishes the core subject; while the \textit{attribute}, say ``NFS Data Transfer Support'', specifies the precise property or characteristic undergoing change. The inclusion of the \textit{concept}, say ``Deployment Infra. and Runtime Support Enhancements'', is critical for contextual disambiguation, ensuring the question targets a specific facet of the entity rather than its broader context. This three-part hierarchical grounding directly leverages our structured knowledge representation to construct meaningful, targeted queries, preventing loosely framed questions that might elicit overly broad responses. An example question: ``What is the maximum data storage capacity supported for NFS data transfer on Snowball Edge devices?'', is shown in \textit{step-5} of Figure~\ref{fig:main-framework-image}.

To ensure high quality of our benchmark, we incorporate a post-generation cleaning step. We remove: (a) questions that elicit exhaustive lists as responses (e.g., ``In which regions are EC2 R6i instances available?''). To correctly answer such questions, we would require a temporally complete \textit{AWS} documentation coverage that is beyond the scope of our input corpus. (b) Any duplicate questions targeting semantically equivalent knowledge.

\textbf{Knowledge Probing Formats:} Previous work~\cite{song2025large} has shown that LLM performance significantly varies across different question types. To understand the impact of knowledge probing format in LLMs' ability to answer questions related to evolving knowledge, we generate three different question types as described below.

\textbf{(1) Open-ended Questions:} Obtained directly using the question generation pipeline explained above. These questions evaluate the LLM's ability to extract the most up-to-date information from its parametric knowledge without any cues.

\textbf{(2) Multiple-Choice Questions (MCQ):} By presenting the correct answer alongside plausible distractors, we can assess an LLM's ability to identify current information when it is explicitly provided in context. For an open-ended question, we construct a corresponding MCQ variant using an LLM. Specifically, by providing the \textit{open-ended question}, \textit{gold answer} (detailed in Section~\ref{subsubsec:gold-ans-curation}) and a list of potential \textit{distractor knowledge chunks} (from related concepts), we prompt an LLM to generate an MCQ question, which probes for the same knowledge as its open-ended counterpart. When generating distractors, we prioritize older and previously valid answers. If historical answers are insufficient in number, we generate plausible but incorrect distractors from related concepts.

\textbf{(3) Verifiable QA:} This format probes the consistency and robustness of an LLM's internal knowledge. For each open-ended question, we generate a pair of ``yes/no'' questions: (a) Yes-eliciting question: presents the correct and current fact and asks for confirmation (e.g., ``Is \textit{90 days} the maximum historical data retention period accessible through Amazon Connect's GetMetricDataV2 API?''). (b) No-eliciting question: presents a plausible but incorrect fact and asks for confirmation (e.g., ``Is \textit{14 days} the maximum historical data retention period accessible through Amazon Connect's GetMetricDataV2 API?'')

\subsubsection{Gold Answer Curation}
\label{subsubsec:gold-ans-curation}
For each question, all documents associated with the chunks within the \textit{attribute} cluster are considered as potential sources for an answer. Our gold answer curation process employs a multi-stage, LLM-guided prompting strategy to ensure accuracy. First, for each potential source document, we prompt an LLM to determine if it contains sufficient and relevant evidence to answer the given question. This initial verification step serves as a critical guardrail, preventing the LLM from attempting to extract answers from irrelevant or insufficient textual segments. Next, based on the LLM's positive validation of evidence sufficiency, we instruct it to extract a precise, well-grounded answer from each identified relevant document. Finally, after obtaining individual answers from all relevant source documents, we use the LLM to synthesize and combine them into a comprehensive gold answer. This step ensures that the final gold answer accurately reflects the information across all available relevant documents for that specific question and timestamp. %

\textbf{Consideration of LLM Knowledge Cut-off Dates:} Our gold answer curation process explicitly accounts for the knowledge cut-off date of the target LLM intended for evaluation. Specifically, during the first step of evidence verification, we filter the potential source documents, discarding any document whose timestamp exceeds the knowledge cut-off date of the LLM being evaluated. For instance, when constructing the benchmark for \texttt{Llama-3.1-8B-Instruct} model (whose knowledge cut-off date is December 2023), all documents published after this date are excluded from the answer curation process for that specific evaluation context. This ensures a fair and accurate evaluation of each LLM's inherent parametric knowledge. We perform a final cleaning of the generated QA pairs. We discard all non-evolving questions, for which only one single answer was identified across all timestamps before the target LLM's knowledge cut-off date. We also discard any questions for which the identified answers show no substantive knowledge change over time.

All the prompt structures used in different steps of our framework are shown in Appendix~\ref{app-subsec:benchmark-creation-related}.

\section{Evolving Knowledge - Question Answering Benchmark (\texttt{evolveQA})}
\label{sec:evolveQA}
Building upon the methodology described in Section~\ref{sec:framework}, we construct \texttt{evolveQA}, a comprehensive QA benchmark specifically designed to test LLMs on temporally evolving knowledge. For its construction, we use \texttt{Claude Sonnet-3.5-v2} for all LLM calls. We acknowledge that, in principle, any sufficiently capable LLM can be substituted. All hyperparameters can be found in Appendix~\ref{app:hyperparameters-preprocessing}.
\vspace{-0.2cm}

\subsection{Knowledge Domains} 
We curate \texttt{evolveQA} from $3$ distinct, time-stamped corpora, chosen to represent varying levels of structure, and domain specificity. Dynamic nature of knowledge in these domains makes them suitable for studying temporal knowledge conflicts. The domains are as follows.

\textbf{AWS What's New Feed:} This corpus comprises publicly available updates and announcements on Amazon Web Services (AWS) from 2019 to 2024.

\textbf{Azure Updates:} Similar in structure to the AWS feed, this corpus covers publicly available updates on Microsoft Azure services. It provides a parallel, yet distinct, technical domain to test LLMs on evolving knowledge. 

\textbf{WHO - Disease Outbreak News (WHO-DONs):} This corpus consists of articles published by the World Health Organization (WHO) regarding global disease and virus outbreaks, spanning from 1997 till 2025. Compared to cloud domains, WHO-DONs represents a significantly different domain. In addition, document length is at least $\approx 110$ words longer than the other two domains with several sections. By breaking down each document into its respective sections, we ingest this structurally diverse data into our framework.

\vspace{-0.1cm}
\subsection{LLMs Used} 
As we discussed in Section \ref{sec:framework}, an LLM's internal knowledge is temporally bounded by its training cut-off date, which directly impacts how these models handle conflicting information without any other supporting mechanism such as Retrieval Augmented Generation (RAG)~\cite{lewis2020retrieval}. As a consequence, we develop \texttt{evolveQA} for a comprehensive suite of $12$ prominent LLMs, each evaluation set aligned with their respective cutoff date. In summary, based on $6$ different knowledge cut-off dates, we selected $6$ open-source and $6$ closed-source models spanning $4$ distinct model families. The knowledge cut-off dates were collected from the model providers' official websites. 

To rigorously validate that the models indeed do not possess knowledge beyond their reported cut-off dates, we conducted an ablation. On our input corpus, we performed a Membership Inference Attack (MIA)~\cite{shi2023detecting} using \texttt{LLama-3.1-8B-instruct}. We found a consistent performance gap in each domain when comparing the relative scores, for the documents \textit{before} and \textit{after} the model's cut-off date. This empirical validation confirms that the chosen input corpus and model cut-off dates are well-aligned with the model's internal knowledge. Appendix~\ref{app:mia-subsec} provides details of this ablation.

\vspace{-0.2cm}
\subsection{Benchmark Quality}
\vspace{-0.1cm}
Prior to evaluating LLMs on \texttt{evolveQA}, we analyze its quality through a human evaluation on a randomly sampled $100$ QA pairs from the benchmark (see Appendix~\ref{app-subsubsec:evolveQA-human-eval} for more details). We observe that the questions from our benchmark are well-grounded to their \{\textit{entity, concept, attribute}\} tuples in $92\%$ of the cases, $94\%$ of the identified attributes evolved over time, and $89\%$ of the questions are of high quality$-$targeting specific non-trivial knowledge. On coupling this with the gold answer correctness of $> 91\%$, we ensure the reliability of our benchmark for LLM evaluation. Table~\ref{tab:evolveQA-statistics} shows \texttt{evolveQA} statistics.

\begin{table}[ht!]
\centering
\resizebox{\columnwidth}{!}{%
\begin{tabular}
{>{\centering\arraybackslash}m{1.25cm}|>{\centering\arraybackslash}m{1.25cm}|>{\arraybackslash}m{7cm}}
\toprule
\textbf{Cutoff Dates} & \textbf{\# of QA pairs} & \textbf{Relevant LLMs based on cutoff} \\
\midrule

08/31/23 & 717 & Claude 3 Haiku~\cite{anthropic2024claude3} \\
\hline
12/31/23 & 775 & Llama 3.1 8B Inst., 3.3 70B Inst.; ~\cite{dubey2024llama} \\
\hline
04/30/24 & 818 & Claude 3.5 Sonnet V2, V1~\cite{anthropic2024claude35} \\
\hline
05/31/24 & 849 & GPT-4.1-mini~\cite{achiam2023gpt}, GPT-5-mini\footnotemark \\
\hline
07/31/24 & 886 & Claude 3.5 Haiku~\cite{anthropic2024claude35} \\
\hline
08/31/24 & 922 & Gemma3 4B, 27B Inst.~\cite{team2025gemma}; Llama4–Scout, Maverick Inst.~\cite{meta2025llama} \\
\bottomrule
\end{tabular}}
\caption{\small \texttt{evolveQA} statistics for different cutoff dates. Full data distribution for each domain is provided in Appendix~\ref{app-subsec:evolveQA-data-distribution-full-table}.}%
\vspace{-4pt}
\label{tab:evolveQA-statistics}
\end{table}
\footnotetext{\tiny\url{https://platform.openai.com/docs/models/gpt-5-mini}}

\vspace{-0.4cm}
\section{Evaluation of LLMs on \texttt{evolveQA}}
\label{sec:experimental-eval}
\vspace{-0.2cm}
\subsection{Experimental Setup}
\textbf{Control Dataset:} To benchmark LLMs' performance on evolving knowledge, we first establish their baseline performance in each domain using a control dataset. We construct a temporally non-conflicting control set by removing the documents from source corpora that are associated with top-$4$ most temporally-conflicting concept-clusters (identified during \texttt{evolveQA} creation), then excluding all documents pertaining to questions in \texttt{evolveQA}, sampling $\approx 1200$ remaining documents, and generating QA pairs targeting static facts. This control dataset enables us to isolate the impact of temporal conflicts and establish each model's baseline performance on static facts within the same domains (see Appendix~\ref{app-subsec:control-set-statistics} for control set statistics).

\textbf{Evaluation Metrics:} We evaluate LLMs' performance in the \texttt{evolveQA} and control datasets using two metrics: \textbf{(a) Accuracy:} For all three question types, this measures the proportion of correct answers. For open-ended QA, we also measure \textit{Outdated Response Rate} (\texttt{ORR}), that measures what proportion of answers are factually correct for a past time period but are outdated based on the most recent knowledge. \textbf{(b) Hard Accuracy for Verifiable QA:} This metric measures the proportion of cases where a model correctly answers both the ``yes-eliciting'' and ``no-eliciting'' variants of the same underlying fact. 

\textbf{Evaluation Methodology:} Exact-match metrics are insufficient to evaluate open-ended questions. Hence, we employ a \textbf{model-as-a-judge} evaluation scheme~\cite{zheng2023judging} for automatic evaluation. In particular, we utilize \texttt{Claude Sonnet-4} as the judge, chosen for its strong instruction-following and nuanced semantic understanding. Along with the \textit{question}, we provide the \textit{LLM-generated response}, the \textit{gold answer} and \textit{answers} from individual documents from previous timestamps to the LLM-judge. Then, we prompt the LLM-judge to validate if the \textit{LLM-generated response} is \textit{CORRECT\_AND\_CURRENT:} indicating that the LLM-generated answer is authoritative; \textit{OUTDATED:} indicating that the LLM-generated answer is aligning with a previously valid answer; \textit{INCORRECT:} indicating that the answer is neither authoritative nor outdated. The evaluation rubric prompt is provided in Appendix~\ref{app-subsec:eval-rubric-model-judge}. We also perform a meta evaluation of the quality of the LLM judge and observe that the judge achieves an accuracy of $\approx 96\%$ while evaluating open-ended responses (details in Appendix~\ref{app-subsec:model-judge-meta-human-eval}).

\subsection{Results and Ablations}
\label{subsec:results-and-ablations}
\textbf{Main results:} The main results are presented in Table~\ref{tab:model-benchmark}. We observe a consistent and significant degradation in performance for all evaluated models on the temporally conflicting \texttt{evolveQA} benchmark compared to the non-conflicting control dataset. In open-ended QA, we observe the most substantial and consistent performance drops across all domains when compared to the performance in the control set. In \textit{AWS domain}, we observe performance drops ranging between $\approx 6\% - 26\%$. It is even more severe in the \textit{WHO-DONs} domain, with reductions between $\approx 8\% - 31\%$, while it is up to $27\%$ in \textit{Azure} domain, when measured across all models. This confirms that even state-of-the-art LLMs struggle to retrieve and prioritize the most current information when faced with evolving facts stored in their parameters.

\begin{table*}[!h]
\small
\centering
\resizebox{2\columnwidth}{!}{%
\begin{tabular}
{>{\centering\arraybackslash}m{1.3cm}|>{\arraybackslash}m{1.7cm}|>{\arraybackslash}m{4.3cm}|>{\centering\arraybackslash}m{1.6cm}|>{\centering\arraybackslash}m{1.6cm}||>{\centering\arraybackslash}m{1.6cm}|>{\centering\arraybackslash}m{1.6cm}||>{\centering\arraybackslash}m{1.8cm}|>{\centering\arraybackslash}m{1.8cm}}
\toprule
& & & \multicolumn{2}{c||}{\textbf{Open-Ended QA (\%)}} & \multicolumn{2}{c||}{\textbf{MCQ (\%)}} & \multicolumn{2}{c}{\textbf{Verifiable QA (\%)}} \\ 
\cmidrule{4-5} \cmidrule{6-7} \cmidrule{8-9}
\textbf{Domains} & \textbf{Model Families} & \textbf{Models} & \textbf{evolveQA Acc.} & \textbf{Control Acc.} & \textbf{evolveQA Acc.} & \textbf{Control Acc.} & \textbf{evolveQA Hard Acc.} & \textbf{Control Hard Acc.} \\
\midrule

\multirow{12}{*}{\textbf{\rotatebox[origin=c]{90}{AWS Whatsnew Feed}}}    
& \multirow{4}{*}{Anthropic} 
& Claude 3.5-sonnet-v2 & 26.95 & 46.39 & 63.33 & 82.49 & 42.64 & 68.03 \\
& & Claude 3.5-sonnet-v1 & 22.9 & 44.41 & 63.33 & 83.47 & 40.94 & 67.05 \\
& & Claude 3 Haiku & 15.93 & 33.84 & 62.47 & 68.07 & 36.99 & 68.39 \\
& & Claude 3.5 Haiku & 22.64 & 42.14 & 67.69 & 73.24 & 37.82 & 43.89 \\
\cmidrule{2-9}
& \multirow{4}{*}{Llama} 
& Llama 3.1-8B-Instruct & 14.44 & 20.33 & 61.84 & 67.53 & 35.76 & 49.04 \\
& & Llama 3.3-70B-Instruct & 18.53 & 33.13 & 64.99 & 73.78 & 36.64 & 72.7 \\
& & Llama 4-Scout-instruct & 17.09 & 31.78 & 68.32 & 75.35 & 41.92 & 65.43 \\
& & Llama 4-Maverick-instruct & 22.18 & 36.45 & 66.3 & 79.51 & 40.04 & 80.53 \\
\cmidrule{2-9}
& \multirow{2}{*}{OpenAI} 
& GPT-4.1-mini & 22.86 & 41.69 & 62.78 & 78.24 & 39.84 & 60 \\
& & GPT-5-mini & 29.62 & 55.4 & 59.96 & 82.08 & 42.71 & 65.71 \\
\cmidrule{2-9}
& \multirow{2}{*}{Gemma} 
& Gemma3-4B-instruct & 12.18 & 18.83 & 52.85 & 58.24 & 35.53 & 25.62 \\
& & Gemma3-27B-instruct & 21.82 & 34.71 & 64.46 & 71.65 & 39.29 & 47.07 \\
\midrule

\multirow{12}{*}{\textbf{\rotatebox[origin=c]{90}{Azure Updates}}}    
& \multirow{4}{*}{Anthropic} 
& Claude 3.5-sonnet-v2 & 22.95 & 49.82 & 70.33 & 77.73 & 49.42 & 58.33 \\
& & Claude 3.5-sonnet-v1 & 21.31 & 46.24 & 70.88 & 78.91 & 47.09 & 55.95 \\
& & Claude 3 Haiku & 18.18 & 27.1 & 70.78 & 64.44 & 40.41 & 58.72 \\
& & Claude 3.5 Haiku & 21.32 & 41.5 & 68.72 & 64.75 & 40.22 & 39.69 \\
\cmidrule{2-9}
& \multirow{4}{*}{Llama} 
& Llama 3.1-8B-Instruct & 17.58 & 16.85 & 63.64 & 60.4 & 45.81 & 43.09 \\
& & Llama 3.3-70B-Instruct & 24.24 & 36.26 & 70.3 & 73.2 & 46.45 & 59.35 \\
& & Llama 4-Scout-instruct & 18.72 & 31.58 & 72.64 & 71.76 & 48.4 & 55.04 \\
& & Llama 4-Maverick-instruct & 23.15 & 37.19 & 76.12 & 75.95 & 47.34 & 75.97 \\
\cmidrule{2-9}
& \multirow{2}{*}{OpenAI} 
& GPT-4.1-mini & 26.84 & 38.35 & 71.81 & 76.56 & 39.89 & 55.95 \\
& & GPT-5-mini & 29.47 & 49.82 & 67.55 & 76.95 & 42.13 & 51.19 \\
\cmidrule{2-9}
& \multirow{2}{*}{Gemma} 
& Gemma3-4B-instruct & 14.29 & 15.09 & 54.73 & 51.53 & 32.98 & 29.46 \\
& & Gemma3-27B-instruct & 25.12 & 30.88 & 69.15 & 64.12 & 46.28 & 40.7 \\
\midrule

\multirow{12}{*}{\textbf{\rotatebox[origin=c]{90}{WHO$-$Disease Outbreak News}}}    
& \multirow{4}{*}{Anthropic} 
& Claude 3.5-sonnet-v2 & 38.41 & 64.09 & 73.1 & 86.19 & 67.67 & 79.31 \\
& & Claude 3.5-sonnet-v1 & 34.23 & 62.43 & 71.72 & 90.61 & 63.91 & 76.44 \\
& & Claude 3 Haiku & 29.41 & 60.69 & 66.67 & 81.5 & 47.9 & 81.93 \\
& & Claude 3.5 Haiku & 33.33 & 63.44 & 72.08 & 80.11 & 58.62 & 76.97 \\
\cmidrule{2-9}
& \multirow{4}{*}{Llama} 
& Llama 3.1-8B-Instruct & 27.4 & 45.71 & 67.36 & 73.14 & 37.5 & 41.67 \\
& & Llama 3.3-70B-Instruct & 39.04 & 59.43 & 68.75 & 83.43 & 53.12 & 72.62 \\
& & Llama 4-Scout-instruct & 31.36 & 55.91 & 68.71 & 82.26 & 56.67 & 75.28 \\
& & Llama 4-Maverick-instruct & 33.14 & 59.14 & 76.07 & 84.95 & 54.19 & 86.45 \\
\cmidrule{2-9}
& \multirow{2}{*}{OpenAI} 
& GPT-4.1-mini & 39.1 & 68.13 & 72.19 & 81.87 & 62.14 & 81.71 \\
& & GPT-5-mini & 50.64 & 71.43 & 72.85 & 86.26 & 67.86 & 86.29 \\
\cmidrule{2-9}
& \multirow{2}{*}{Gemma} 
& Gemma3-4B-instruct & 23.67 & 31.72 & 63.19 & 64.52 & 41.94 & 19.1 \\
& & Gemma3-27B-instruct & 39.05 & 63.98 & 71.17 & 79.51 & 50.97 & 61.24 \\
\bottomrule

\end{tabular}}
\caption{\small Performance comparison of $4$ model families across $3$ data domains. Results are reported for $3$ different knowledge probing formats$-$ for both \texttt{evolveQA} and non-conflicting control dataset. Results are based on a single execution of the LLM.}
\label{tab:model-benchmark}
\vspace{-0.6cm} 
\end{table*}

\textbf{Impact of question format:} We observe that performance varies across question types. Accuracy improves when the question is framed as an MCQ question or a verifiable question, when compared to open-ended question format (Table~\ref{tab:model-benchmark}). Specifically, we observe accuracy improvements of $\approx 22\% - 54\%$ in MCQ format and $\approx 10\% - 30\%$ in verifiable QA format, when compared against the open-ended questions in \texttt{evolveQA}. It indicates that presence of the correct information in the context either in MCQ format or in the verifiable QA format, helps the LLMs select the correct answer. However, in both of the question types, performance lags significantly when compared to the control dataset. This finding also verifies that the LLMs have the most up-to-date information in their parametric knowledge, however, fails to recall them when probed in an open-ended setting.

\textbf{Notable Model-Specific Anomalies:} In the Azure domain, for a few LLMs (e.g., Claude Haiku, Llama 3), MCQ performance on \texttt{evolveQA} is comparable or even slightly higher than the performance in the control dataset. Upon deep dive, we identify $17$ instances in the non-conflicting control dataset where \texttt{Claude-3.5-Haiku} answered correctly in the open-ended setting but failed in its corresponding MCQ format (examples provided in Appendix~\ref{app-subsubsec:notable-anomalies}), which contributed to its lower accuracy in the control dataset. Surprisingly, we do not observe this behavior on \texttt{evolveQA} with \texttt{Haiku}. With \texttt{Llama-3.1-8B model}, we observe a comparatively minor gap in performance between the control dataset and \texttt{evolveQA}. This can be attributed to its lower performance in general, as seen in open-ended QA format for \textit{Azure} and \textit{AWS Domains}. In verifiable QA control set, we observe that \texttt{Gemma-4B} is generally biased to responding ``yes'', agreeing with $\approx 81\%$ of ``yes''-eliciting questions but also incorrectly agreeing with $\approx 63\%$ of ``no''-eliciting questions (averaged across the $3$ domains). Since \texttt{Gemma-4B} fails to correctly answer both the ``yes-no'' questions, its \textit{Hard Acc.} performance is lower across the $3$ domains in the control dataset.

\textbf{Outdated response rate (ORR):} LLMs provide outdated responses more frequently than completely incorrect ones in open-ended QA, with \texttt{ORR} ranging between $\approx 23\% - 54\%$ (as detailed in Appendix~\ref{app-subsec:orr}). Interestingly, we find that the knowledge of the correct answer is often latently present, even when an outdated response is generated. We test this by examining instances where a model failed in an open-ended question format for providing an outdated answer and then assessing its performance on its corresponding MCQ format. In a substantial number of cases, the model selected the valid option, indicating that the model is aware of the authoritative response. Averaged across all models, this behavior occurred $32.44\%$ in \textit{WHO-DONs} domain, $41.34\%$ in \textit{Azure} domain and $45.2\%$ in \textit{AWS} domain.

\textbf{Impact of explicit context in prompt:} We investigate whether providing more context related to the question helps LLMs handle temporal conflicts in open-ended QA by testing three prompt variations: (1) providing ground-truth Entity-Concept-Attribute (E-C-A) from our benchmark in prompt; (2) providing ``current date'' in the prompt; (3) providing both E-C-A and ``current date''. Results show that providing only the ``current date'' generally does not improve LLMs ability to recall authoritative information. However, providing E-C-A context generally improves performance, providing gains up to $11\%$ (see Appendix~\ref{app-subsec:eca-context} for details).

\textbf{Impact of superseding knowledge:} Conflicts in \texttt{evolveQA} can be categorized into $2$ types: (1) \textit{additive knowledge} that complements existing facts (e.g., ``How does Kendra access user information?'' has multiple valid answers); (2) \textit{superseding knowledge} that invalidates previous facts (eg., ``How many AWS regions support Kendra?'' has only one valid answer). To analyze the more challenging superseding knowledge cases, we isolate MCQ instances where outdated facts (previously valid) serve as distractors. Compared with the entire benchmark, we observe an increased vulnerability to superseding knowledge across models. In MCQ format, models show significant degradation (e.g., $~15\%$ drop for \texttt{GPT-4.1-mini} in \textit{AWS} and \textit{WHO-DONs}, similar for \texttt{GPT-5-mini} in \textit{Azure}), suggesting strong interference from historically correct distractors. In open-ended questions, performance drops up to $9\%$ in certain cases. Notably, GPT family models consistently show higher degradation across both formats, indicating potential architectural sensitivity to factual updates (details in Appendix~\ref{app-subsec:superseding-knowledge}).

\vspace{-0.2cm}
\section{Conclusion}
\vspace{-0.185cm}
In this paper, we introduce \texttt{evolveQA} to evaluate LLMs on temporal knowledge conflicts using naturally evolving real-world data. Our findings reveal that while LLMs contain updated knowledge in their parameters, they struggle to retrieve it reliably, particularly in open-ended queries. This work exposes critical limitations of modern LLMs and provides a benchmark for developing temporally robust LLMs in dynamically evolving domains.

\section*{Limitations}

\textbf{Availability of temporal information:} Our methodology relies on availability of reliable, timestamped documents. The entire pipeline, from identifying evolving facts to curating time-sensitive gold answers, relies on this metadata. Consequently, our framework cannot be directly applied to corpora lacking explicit chronological information.

\textbf{Limitation in quantifying Recall:} A key methodological challenge is quantifying the recall of our fact identification process. The absence of a comprehensive ground-truth dataset for evolving facts within unstructured corpora is an inherent limitation of this task. Consequently, while we can evaluate the precision of the facts we identify, we cannot measure the completeness of our framework in capturing every possible instance of evolving knowledge. For a similar reason, during the construction of the control dataset, we cannot quantify the number of cases where we successfully identify static knowledge.

\textbf{Modules and languages studied:} Our approach is limited to the text module in the English language only, while evolving knowledge in other modules (e.g., image, table, and so on) and languages can also be explored. We leave it as a future work.

\textbf{Dependency on backbone LLM's capability:} Our approach depends on a number of LLM calls in order to leverage their reasoning capability in different stages of the proposed algorithm. In our experiment, we use \texttt{Sonnet-3.5} as the backbone LLM, however, we note that any other reasonably capable LLM can be used for the prompting steps. We leave the correlation study between the quality of the resulting benchmark and the capability of the backbone LLM, as a future work.

\textbf{Fixed cost for benchmark construction:} To ensure high quality of our benchmark, our LLM-based approach uses \texttt{Claude Sonnet 3.5} throughout the pipeline. This results in a fixed cost of roughly $400$ USD for constructing \texttt{evolveQA}. We note that this is a lot cheaper than relying on human annotations to evaluate LLMs on temporally conflicting data from three domains.

\textbf{Knowledge conflict mitigation:} Our study focuses on understanding how LLMs handle conflicting information stored in their parametric knowledge. While some studies have suggested using retrieval-augmented generation (RAG) to address these conflicts, our goal is different. Instead of trying to fix these conflicts, we aim to understand how LLMs behave when they encounter contradictory information. We hope this understanding will help develop solutions during model training or through other post-training methods in the future.

\textbf{Limited temporal scope of the input corpus:} Our study relies on datasets with limited temporal scope. For instance, \textit{WHO-DONs} covers updates from 1997$-$2024. As a result, our experiments are restricted to question types for which we have sufficient evidence within this period. In particular, we cannot have questions requiring  \textit{exhaustive} or \textit{comprehensive} coverage. Questions such as ``In which regions did HCV outbreak occur?'' cannot be fully answered or automatically evaluated without access to a temporally \textit{complete} source documentation. This limitation stems from the unavailability of complete knowledge in the domains in the corresponding websites.

\section*{Ethical Considerations}\label{sec:ethics}
To the best of our knowledge, we did not violate any ethical code while conducting the research work described in this paper. We reported all implementation and dataset details in the paper for reproducibility and upon acceptance we will release our codes and datasets for future research. The datasets (from public websites) used in this paper are publicly available and permitted for scientific research. The human evaluation details are presented in Appendix \ref{subsec:app-human-eval}. We perform extensive ablation studies and meta evaluation of the model judge used in the work, in order to provide the readers an idea about potential error patterns and risks associated with our proposed method and reported results.

\bibliography{custom}

\begin{thebibliography}{44}
\providecommand{\natexlab}[1]{#1}

\bibitem[{Achiam et~al.(2023)Achiam, Adler, Agarwal, Ahmad, Akkaya, Aleman, Almeida, Altenschmidt, Altman, Anadkat et~al.}]{achiam2023gpt}
Josh Achiam, Steven Adler, Sandhini Agarwal, Lama Ahmad, Ilge Akkaya, Florencia~Leoni Aleman, Diogo Almeida, Janko Altenschmidt, Sam Altman, Shyamal Anadkat, and 1 others. 2023.
\newblock Gpt-4 technical report.
\newblock \emph{arXiv preprint arXiv:2303.08774}.

\bibitem[{Anthropic(2024{\natexlab{a}})}]{anthropic2024claude3}
AI~Anthropic. 2024{\natexlab{a}}.
\newblock The claude 3 model family: Opus, sonnet, haiku.
\newblock \emph{Claude-3 Model Card}, 1(1):4.

\bibitem[{Anthropic(2024{\natexlab{b}})}]{anthropic2024claude35}
AI~Anthropic. 2024{\natexlab{b}}.
\newblock Claude 3.5 sonnet model card addendum.
\newblock \emph{Claude-3.5 Model Card}, 3(6).

\bibitem[{Brown et~al.(2020)Brown, Mann, Ryder, Subbiah, Kaplan, Dhariwal, Neelakantan, Shyam, Sastry, Askell et~al.}]{brown2020language}
Tom Brown, Benjamin Mann, Nick Ryder, Melanie Subbiah, Jared~D Kaplan, Prafulla Dhariwal, Arvind Neelakantan, Pranav Shyam, Girish Sastry, Amanda Askell, and 1 others. 2020.
\newblock Language models are few-shot learners.
\newblock \emph{Advances in neural information processing systems}, 33:1877--1901.

\bibitem[{Campello et~al.(2013)Campello, Moulavi, and Sander}]{campello2013density}
Ricardo~JGB Campello, Davoud Moulavi, and J{\"o}rg Sander. 2013.
\newblock Density-based clustering based on hierarchical density estimates.
\newblock In \emph{Pacific-Asia conference on knowledge discovery and data mining}, pages 160--172. Springer.

\bibitem[{Chang et~al.(2024)Chang, Wang, Wang, Wu, Yang, Zhu, Chen, Yi, Wang, Wang et~al.}]{chang2024survey}
Yupeng Chang, Xu~Wang, Jindong Wang, Yuan Wu, Linyi Yang, Kaijie Zhu, Hao Chen, Xiaoyuan Yi, Cunxiang Wang, Yidong Wang, and 1 others. 2024.
\newblock A survey on evaluation of large language models.
\newblock \emph{ACM transactions on intelligent systems and technology}, 15(3):1--45.

\bibitem[{Chen et~al.(2022)Chen, Zhang, and Choi}]{chen2022rich}
Hung-Ting Chen, Michael~JQ Zhang, and Eunsol Choi. 2022.
\newblock Rich knowledge sources bring complex knowledge conflicts: Recalibrating models to reflect conflicting evidence.
\newblock \emph{arXiv preprint arXiv:2210.13701}.

\bibitem[{Chen et~al.(2023)Chen, Bruno, and Roth}]{chen2023towards}
Sihao Chen, William Bruno, and Dan Roth. 2023.
\newblock Towards corpus-scale discovery of selection biases in news coverage: Comparing what sources say about entities as a start.
\newblock \emph{arXiv preprint arXiv:2304.03414}.

\bibitem[{Dhingra et~al.(2022)Dhingra, Cole, Eisenschlos, Gillick, Eisenstein, and Cohen}]{dhingra2022time}
Bhuwan Dhingra, Jeremy~R Cole, Julian~Martin Eisenschlos, Daniel Gillick, Jacob Eisenstein, and William~W Cohen. 2022.
\newblock Time-aware language models as temporal knowledge bases.
\newblock \emph{Transactions of the Association for Computational Linguistics}, 10:257--273.

\bibitem[{Dong et~al.(2023)Dong, Xu, Kong, Sui, and Li}]{dong2023statistical}
Qingxiu Dong, Jingjing Xu, Lingpeng Kong, Zhifang Sui, and Lei Li. 2023.
\newblock Statistical knowledge assessment for large language models.
\newblock \emph{Advances in Neural Information Processing Systems}, 36:29812--29830.

\bibitem[{Dubey et~al.(2024)Dubey, Jauhri, Pandey, Kadian, Al-Dahle, Letman, Mathur, Schelten, Yang, Fan et~al.}]{dubey2024llama}
Abhimanyu Dubey, Abhinav Jauhri, Abhinav Pandey, Abhishek Kadian, Ahmad Al-Dahle, Aiesha Letman, Akhil Mathur, Alan Schelten, Amy Yang, Angela Fan, and 1 others. 2024.
\newblock The llama 3 herd of models.
\newblock \emph{arXiv e-prints}, pages arXiv--2407.

\bibitem[{Elazar et~al.(2021)Elazar, Kassner, Ravfogel, Ravichander, Hovy, Sch{\"u}tze, and Goldberg}]{elazar-etal-2021-measuring}
Yanai Elazar, Nora Kassner, Shauli Ravfogel, Abhilasha Ravichander, Eduard Hovy, Hinrich Sch{\"u}tze, and Yoav Goldberg. 2021.
\newblock \href {https://doi.org/10.1162/tacl_a_00410} {Measuring and improving consistency in pretrained language models}.
\newblock \emph{Transactions of the Association for Computational Linguistics}, 9:1012--1031.

\bibitem[{Elsahar et~al.(2018)Elsahar, Vougiouklis, Remaci, Gravier, Hare, Laforest, and Simperl}]{elsahar-etal-2018-rex}
Hady Elsahar, Pavlos Vougiouklis, Arslen Remaci, Christophe Gravier, Jonathon Hare, Frederique Laforest, and Elena Simperl. 2018.
\newblock \href {https://aclanthology.org/L18-1544/} {{T}-{RE}x: A large scale alignment of natural language with knowledge base triples}.
\newblock In \emph{Proceedings of the Eleventh International Conference on Language Resources and Evaluation ({LREC} 2018)}, Miyazaki, Japan. European Language Resources Association (ELRA).

\bibitem[{Fatemi et~al.(2024)Fatemi, Kazemi, Tsitsulin, Malkan, Yim, Palowitch, Seo, Halcrow, and Perozzi}]{fatemi2024test}
Bahare Fatemi, Mehran Kazemi, Anton Tsitsulin, Karishma Malkan, Jinyeong Yim, John Palowitch, Sungyong Seo, Jonathan Halcrow, and Bryan Perozzi. 2024.
\newblock Test of time: A benchmark for evaluating llms on temporal reasoning.
\newblock \emph{arXiv preprint arXiv:2406.09170}.

\bibitem[{Kaddour et~al.(2023)Kaddour, Harris, Mozes, Bradley, Raileanu, and McHardy}]{kaddour2023challenges}
Jean Kaddour, Joshua Harris, Maximilian Mozes, Herbie Bradley, Roberta Raileanu, and Robert McHardy. 2023.
\newblock Challenges and applications of large language models.
\newblock \emph{arXiv preprint arXiv:2307.10169}.

\bibitem[{Kim et~al.(2023)Kim, Yoon, Ye, Bae, Ho, Hwang, and Yun}]{kim2023carpe}
Yujin Kim, Jaehong Yoon, Seonghyeon Ye, Sangmin Bae, Namgyu Ho, Sung~Ju Hwang, and Se-Young Yun. 2023.
\newblock Carpe diem: On the evaluation of world knowledge in lifelong language models.
\newblock \emph{arXiv preprint arXiv:2311.08106}.

\bibitem[{Kortukov et~al.(2024)Kortukov, Rubinstein, Nguyen, and Oh}]{kortukov2024studying}
Evgenii Kortukov, Alexander Rubinstein, Elisa Nguyen, and Seong~Joon Oh. 2024.
\newblock Studying large language model behaviors under context-memory conflicts with real documents.
\newblock \emph{arXiv preprint arXiv:2404.16032}.

\bibitem[{Lewis et~al.(2020)Lewis, Perez, Piktus, Petroni, Karpukhin, Goyal, K{\"u}ttler, Lewis, Yih, Rockt{\"a}schel et~al.}]{lewis2020retrieval}
Patrick Lewis, Ethan Perez, Aleksandra Piktus, Fabio Petroni, Vladimir Karpukhin, Naman Goyal, Heinrich K{\"u}ttler, Mike Lewis, Wen-tau Yih, Tim Rockt{\"a}schel, and 1 others. 2020.
\newblock Retrieval-augmented generation for knowledge-intensive nlp tasks.
\newblock \emph{Advances in neural information processing systems}, 33:9459--9474.

\bibitem[{Mallen et~al.(2022)Mallen, Asai, Zhong, Das, Khashabi, and Hajishirzi}]{mallen2022not}
Alex Mallen, Akari Asai, Victor Zhong, Rajarshi Das, Daniel Khashabi, and Hannaneh Hajishirzi. 2022.
\newblock When not to trust language models: Investigating effectiveness of parametric and non-parametric memories.
\newblock \emph{arXiv preprint arXiv:2212.10511}.

\bibitem[{Marjanovi{\'c} et~al.(2024)Marjanovi{\'c}, Yu, Atanasova, Maistro, Lioma, and Augenstein}]{marjanovic2024dynamicqa}
Sara~Vera Marjanovi{\'c}, Haeun Yu, Pepa Atanasova, Maria Maistro, Christina Lioma, and Isabelle Augenstein. 2024.
\newblock Dynamicqa: Tracing internal knowledge conflicts in language models.
\newblock \emph{arXiv preprint arXiv:2407.17023}.

\bibitem[{McInnes et~al.(2018)McInnes, Healy, and Melville}]{mcinnes2018umap}
Leland McInnes, John Healy, and James Melville. 2018.
\newblock Umap: Uniform manifold approximation and projection for dimension reduction.
\newblock \emph{arXiv preprint arXiv:1802.03426}.

\bibitem[{Meta(2025)}]{meta2025llama}
AI~Meta. 2025.
\newblock The llama 4 herd: The beginning of a new era of natively multimodal ai innovation.
\newblock \emph{https://ai. meta. com/blog/llama-4-multimodal-intelligence/, checked on}, 4(7):2025.

\bibitem[{Nakshatri et~al.(2023)Nakshatri, Liu, Chen, Roth, Goldwasser, and Hopkins}]{nakshatri2023using}
Nishanth Nakshatri, Siyi Liu, Sihao Chen, Dan Roth, Dan Goldwasser, and Daniel Hopkins. 2023.
\newblock Using llm for improving key event discovery: Temporal-guided news stream clustering with event summaries.
\newblock In \emph{Findings of the Association for Computational Linguistics: EMNLP 2023}, pages 4162--4173.

\bibitem[{Nakshatri et~al.(2025)Nakshatri, Mehta, Liu, Chen, Hopkins, Roth, and Goldwasser}]{nakshatri2025talking}
Nishanth Nakshatri, Nikhil Mehta, Siyi Liu, Sihao Chen, Daniel~J Hopkins, Dan Roth, and Dan Goldwasser. 2025.
\newblock Talking point based ideological discourse analysis in news events.
\newblock \emph{arXiv preprint arXiv:2504.07400}.

\bibitem[{Nanni et~al.(2017)Nanni, Ponzetto, and Dietz}]{nanni2017building}
Federico Nanni, Simone~Paolo Ponzetto, and Laura Dietz. 2017.
\newblock Building entity-centric event collections.
\newblock In \emph{2017 ACM/IEEE Joint Conference on Digital Libraries (JCDL)}, pages 1--10. IEEE.

\bibitem[{Neeman et~al.(2022)Neeman, Aharoni, Honovich, Choshen, Szpektor, and Abend}]{neeman2022disentqa}
Ella Neeman, Roee Aharoni, Or~Honovich, Leshem Choshen, Idan Szpektor, and Omri Abend. 2022.
\newblock Disentqa: Disentangling parametric and contextual knowledge with counterfactual question answering.
\newblock \emph{arXiv preprint arXiv:2211.05655}.

\bibitem[{{\"O}zer and Y{\i}ld{\i}z(2025)}]{ozer2025question}
Atahan {\"O}zer and {\c{C}}a{\u{g}}atay Y{\i}ld{\i}z. 2025.
\newblock Question answering under temporal conflict: Evaluating and organizing evolving knowledge with llms.
\newblock \emph{arXiv preprint arXiv:2506.07270}.

\bibitem[{Pan et~al.(2021)Pan, Chen, Kan, and Wang}]{pan2021attacking}
Liangming Pan, Wenhu Chen, Min-Yen Kan, and William~Yang Wang. 2021.
\newblock Attacking open-domain question answering by injecting misinformation.
\newblock \emph{arXiv preprint arXiv:2110.07803}.

\bibitem[{Shi et~al.(2023)Shi, Ajith, Xia, Huang, Liu, Blevins, Chen, and Zettlemoyer}]{shi2023detecting}
Weijia Shi, Anirudh Ajith, Mengzhou Xia, Yangsibo Huang, Daogao Liu, Terra Blevins, Danqi Chen, and Luke Zettlemoyer. 2023.
\newblock Detecting pretraining data from large language models.
\newblock \emph{arXiv preprint arXiv:2310.16789}.

\bibitem[{Song et~al.(2025)Song, Chakraborty, Li, and Tavanapong}]{song2025large}
Seok~Hwan Song, Mohna Chakraborty, Qi~Li, and Wallapak Tavanapong. 2025.
\newblock Is large language model performance on reasoning tasks impacted by different ways questions are asked?
\newblock \emph{arXiv preprint arXiv:2507.15707}.

\bibitem[{Su et~al.(2024{\natexlab{a}})Su, Li, Zhang, Zhu, Qu, Zhou, Bowen, Cheng et~al.}]{su2024living}
Zhaochen Su, Juntao Li, Jun Zhang, Tong Zhu, Xiaoye Qu, Pan Zhou, Yan Bowen, Yu~Cheng, and 1 others. 2024{\natexlab{a}}.
\newblock Living in the moment: Can large language models grasp co-temporal reasoning?
\newblock \emph{arXiv preprint arXiv:2406.09072}.

\bibitem[{Su et~al.(2024{\natexlab{b}})Su, Zhang, Qu, Zhu, Li, Sun, Li, Zhang, and Cheng}]{su2024conflictbank}
Zhaochen Su, Jun Zhang, Xiaoye Qu, Tong Zhu, Yanshu Li, Jiashuo Sun, Juntao Li, Min Zhang, and Yu~Cheng. 2024{\natexlab{b}}.
\newblock Conflictbank: A benchmark for evaluating the influence of knowledge conflicts in llm.
\newblock \emph{arXiv preprint arXiv:2408.12076}.

\bibitem[{Tamkin et~al.(2024)Tamkin, McCain, Handa, Durmus, Lovitt, Rathi, Huang, Mountfield, Hong, Ritchie et~al.}]{tamkin2024clio}
Alex Tamkin, Miles McCain, Kunal Handa, Esin Durmus, Liane Lovitt, Ankur Rathi, Saffron Huang, Alfred Mountfield, Jerry Hong, Stuart Ritchie, and 1 others. 2024.
\newblock Clio: Privacy-preserving insights into real-world ai use.
\newblock \emph{arXiv preprint arXiv:2412.13678}.

\bibitem[{Tan et~al.(2023)Tan, Dwivedi-Yu, Li, Mathias, Saeidi, Yan, and Halevy}]{tan2023timelineqa}
Wang-Chiew Tan, Jane Dwivedi-Yu, Yuliang Li, Lambert Mathias, Marzieh Saeidi, Jing~Nathan Yan, and Alon~Y Halevy. 2023.
\newblock Timelineqa: A benchmark for question answering over timelines.
\newblock \emph{arXiv preprint arXiv:2306.01069}.

\bibitem[{Tang et~al.(2025)Tang, Cao, Deng, Ying, Wang, Yang, Zhao, Zhang, Huang, Jiang, and Liao}]{tang-etal-2025-evowiki}
Wei Tang, Yixin Cao, Yang Deng, Jiahao Ying, Bo~Wang, Yizhe Yang, Yuyue Zhao, Qi~Zhang, Xuanjing Huang, Yu-Gang Jiang, and Yong Liao. 2025.
\newblock \href {https://doi.org/10.18653/v1/2025.acl-long.47} {{E}vo{W}iki: Evaluating {LLM}s on evolving knowledge}.
\newblock In \emph{Proceedings of the 63rd Annual Meeting of the Association for Computational Linguistics (Volume 1: Long Papers)}, pages 948--964, Vienna, Austria. Association for Computational Linguistics.

\bibitem[{Team et~al.(2025)Team, Kamath, Ferret, Pathak, Vieillard, Merhej, Perrin, Matejovicova, Ram{\'e}, Rivi{\`e}re et~al.}]{team2025gemma}
Gemma Team, Aishwarya Kamath, Johan Ferret, Shreya Pathak, Nino Vieillard, Ramona Merhej, Sarah Perrin, Tatiana Matejovicova, Alexandre Ram{\'e}, Morgane Rivi{\`e}re, and 1 others. 2025.
\newblock Gemma 3 technical report.
\newblock \emph{arXiv preprint arXiv:2503.19786}.

\bibitem[{Wang et~al.(2025)Wang, Wan, Sun, Chen, and Arık}]{wang2025astuteragovercomingimperfect}
Fei Wang, Xingchen Wan, Ruoxi Sun, Jiefeng Chen, and Sercan~Ö. Arık. 2025.
\newblock Astute rag: Overcoming imperfect retrieval augmentation and knowledge conflicts for large language models.
\newblock \emph{https://arxiv.org/abs/2410.07176}.

\bibitem[{Wang et~al.(2023)Wang, Feng, Wang, Shi, Balachandran, He, and Tsvetkov}]{wang2023resolving}
Yike Wang, Shangbin Feng, Heng Wang, Weijia Shi, Vidhisha Balachandran, Tianxing He, and Yulia Tsvetkov. 2023.
\newblock Resolving knowledge conflicts in large language models.
\newblock \emph{arXiv preprint arXiv:2310.00935}.

\bibitem[{Wei et~al.(2021)Wei, Bosma, Zhao, Guu, Yu, Lester, Du, Dai, and Le}]{wei2021finetuned}
Jason Wei, Maarten Bosma, Vincent~Y Zhao, Kelvin Guu, Adams~Wei Yu, Brian Lester, Nan Du, Andrew~M Dai, and Quoc~V Le. 2021.
\newblock Finetuned language models are zero-shot learners.
\newblock \emph{arXiv preprint arXiv:2109.01652}.

\bibitem[{Xie et~al.(2023)Xie, Zhang, Chen, Lou, and Su}]{xie2023adaptive}
Jian Xie, Kai Zhang, Jiangjie Chen, Renze Lou, and Yu~Su. 2023.
\newblock Adaptive chameleon or stubborn sloth: Revealing the behavior of large language models in knowledge conflicts.
\newblock In \emph{The Twelfth International Conference on Learning Representations}.

\bibitem[{Zhang et~al.(2023)Zhang, Wang, and Shang}]{zhang2023clusterllm}
Yuwei Zhang, Zihan Wang, and Jingbo Shang. 2023.
\newblock Clusterllm: Large language models as a guide for text clustering.
\newblock \emph{arXiv preprint arXiv:2305.14871}.

\bibitem[{Zhao et~al.(2023)Zhao, Yan, Sun, Xing, Meng, Wang, Cheng, Ren, and Yin}]{zhao2023knowing}
Yukun Zhao, Lingyong Yan, Weiwei Sun, Guoliang Xing, Chong Meng, Shuaiqiang Wang, Zhicong Cheng, Zhaochun Ren, and Dawei Yin. 2023.
\newblock Knowing what llms do not know: A simple yet effective self-detection method.
\newblock \emph{arXiv preprint arXiv:2310.17918}.

\bibitem[{Zheng et~al.(2023)Zheng, Chiang, Sheng, Zhuang, Wu, Zhuang, Lin, Li, Li, Xing et~al.}]{zheng2023judging}
Lianmin Zheng, Wei-Lin Chiang, Ying Sheng, Siyuan Zhuang, Zhanghao Wu, Yonghao Zhuang, Zi~Lin, Zhuohan Li, Dacheng Li, Eric Xing, and 1 others. 2023.
\newblock Judging llm-as-a-judge with mt-bench and chatbot arena.
\newblock \emph{Advances in neural information processing systems}, 36:46595--46623.

\bibitem[{Zhou et~al.(2023)Zhou, Liu, Xu, Iyer, Sun, Mao, Ma, Efrat, Yu, Yu et~al.}]{zhou2023lima}
Chunting Zhou, Pengfei Liu, Puxin Xu, Srinivasan Iyer, Jiao Sun, Yuning Mao, Xuezhe Ma, Avia Efrat, Ping Yu, Lili Yu, and 1 others. 2023.
\newblock Lima: Less is more for alignment.
\newblock \emph{Advances in Neural Information Processing Systems}, 36:55006--55021.

\end{thebibliography}

\appendix

\section{Related Work}
\label{app-sec:related-work}

\textbf{Knowledge Conflict in LLMs.}
Understanding the effect of different types of knowledge conflicts in LLMs has recently gained much interest. For example, conflicts in retrieved context~\cite{pan2021attacking, chen2022rich, wang2025astuteragovercomingimperfect}, conflicts between external context and LLMs' internal parametric knowledge~\cite{neeman2022disentqa, wang2023resolving, xie2023adaptive, kortukov2024studying}, and intra-memory conflicts within an LLM's parameters~\cite{elsahar-etal-2018-rex, elazar-etal-2021-measuring, dong2023statistical, zhao2023knowing}. Our work addresses this third category, focusing specifically on temporal knowledge conflicts that arise from outdated information in pre-training corpora~\cite{kaddour2023challenges, zhou2023lima}. Existing works focused on structured knowledge bases like \textit{Wikidata}~\cite{kim2023carpe, marjanovic2024dynamicqa, su2024conflictbank, ozer2025question}; a few relied on Subject-Relation-Object (SRO) triplet extraction-based approaches~\cite{ozer2025question, tang-etal-2025-evowiki} and a few others relied on artificial construction of temporal conflicts by altering timestamps~\cite{su2024conflictbank}. Most works created static benchmarks by ignoring to account for the knowledge cut-off date of the LLMs. While~\cite{tang-etal-2025-evowiki} considers the cut-off date of the models, it is still based on \textit{Wikidata}, where the availability of pre-defined data snapshots at different timestamps aids in the identification of evolving knowledge. Such metadata is not universally present across all corpora. We focus on unstructured, time-stamped corpora in specialized domains, while explicitly accounting for each LLM's knowledge cut-off date.

\textbf{Temporal Reasoning.}
We distinguish our focus on temporal knowledge conflicts from the broader field of temporal reasoning where the focus is on testing LLMs' ability to understand basic temporal concepts such as ordering events chronologically, determining how long events last, and understanding temporal relationships among events~\cite{dhingra2022time, tan2023timelineqa, fatemi2024test}. In contrast, our work focuses on the (in)consistency of factual knowledge stored within an LLM's parameters, as that knowledge evolves over time.

\section{\texttt{evolveQA}}

\subsection{Data Distribution - Across domains and Knowledge Probing Formats}
\label{app-subsec:evolveQA-data-distribution-full-table}
Table~\ref{tab-app:evolveQA_distribution} provides the complete distribution of \texttt{evolveQA} across the three domains of interest$-$ \{\textit{AWS Feed, Azure, WHO-Disease Outbreak News (WHO-DONs)}\}; It also provides the distribution for each knowledge probing format along with the relevant LLMs associated with each knowledge cut-off date.

\begin{table*}[ht!]
\centering
\resizebox{\textwidth}{!}{%
\begin{tabular}{l|ccc|ccc|ccc|c|l}
\toprule
\multicolumn{1}{c|}{} & \multicolumn{9}{c|}{\textbf{No. of QA Pairs Across Different Knowledge Probing Formats (evolveQA)}} & & \multicolumn{1}{c}{} \\
\cmidrule(lr){2-10}
\textbf{Cutoff Date} & \multicolumn{3}{c|}{\textbf{AWS Feed}} & \multicolumn{3}{c|}{\textbf{Azure Updates}} & \multicolumn{3}{c|}{\textbf{WHO-DONs}} & \textbf{Total No. of } & \textbf{Relevant LLMs based on cutoff} \\
\cmidrule(lr){2-4} \cmidrule(lr){5-7} \cmidrule(lr){8-10}
& \textbf{Open-ended} & \textbf{MCQ} & \textbf{Ver. QA} & \textbf{Open-ended} & \textbf{MCQ} & \textbf{Ver. QA} & \textbf{Open-ended} & \textbf{MCQ} & \textbf{Ver. QA} & \textbf{Open-Ended QA Pairs} & \\
\midrule
8/31/23 & 427 & 421 & 419 & 154 & 154 & 146 & 136 & 132 & 119 & 717 & Claude 3 Haiku \\
12/31/23 & 464 & 457 & 453 & 165 & 165 & 155 & 146 & 144 & 128 & 775 & Llama 3.1-8B-inst, 3.3-70B inst. \\
4/30/24 & 486 & 480 & 469 & 183 & 182 & 172 & 149 & 145 & 133 & 818 & Claude 3.5 Sonnet V2, V1 \\
5/31/24 & 503 & 497 & 487 & 190 & 188 & 178 & 156 & 151 & 140 & 849 & GPT-4.1-mini, GPT-5-mini \\
7/31/24 & 530 & 523 & 513 & 197 & 195 & 180 & 159 & 154 & 145 & 886 & Claude 3.5 Haiku \\
8/31/24 & 550 & 543 & 532 & 203 & 201 & 188 & 169 & 163 & 155 & 922 & Gemma3 4B, 27B inst, Llama4-Scout, Maverick inst. \\
\bottomrule
\end{tabular}%
}
\caption{\small Shows a detailed breakdown of \texttt{evolveQA} (open-ended question format) across the three domains - \{\textit{AWS, Azure, and WHO-DONs}\} domain. The table also provides the breakdown with respect to different knowledge probing formats across the three domains. Along with these, it shows the relevant LLMs intended for evaluation for each knowledge cut-off date.}
\label{tab-app:evolveQA_distribution}
\end{table*}

\subsection{Control Set - Data Distribution}
\label{app-subsec:control-set-statistics}
Table~\ref{tab:control_set_distribution} shows the detailed statistics for each knowledge cut-off date and across each domain under consideration, which is kept consistent with \texttt{evolveQA} for a fair comparison.

\begin{table}[ht!]
\small 
\centering
\resizebox{\columnwidth}{!}{%
\begin{tabular}{l|ccc|c}
\toprule
\multirow{2}{*}{\textbf{Cutoff Date}} & \multicolumn{3}{c|}{\textbf{No. of QA pairs}} & \multirow{2}{*}{\textbf{Total No. QA Pairs}} \\
\cmidrule(lr){2-4}
& \textbf{AWS Feed} & \textbf{Azure Updates} & \textbf{WHO-DONs} & \\
\midrule
8/31/23 & 597 & 262 & 173 & 1032 \\
12/31/23 & 664 & 273 & 175 & 1112 \\
4/30/24 & 707 & 279 & 181 & 1167 \\
5/31/24 & 722 & 279 & 182 & 1183 \\
7/31/24 & 738 & 284 & 186 & 1209 \\
8/31/24 & 749 & 285 & 186 & 1220 \\
\bottomrule
\end{tabular}%
}
\caption{\small Shows the data distribution for the control set across the three domains for each knowledge cut-off date (consistent with \texttt{evolveQA} for a fair comparison.}
\label{tab:control_set_distribution}
\end{table}

\subsection{Hyperparameters / Implementation Details}
\label{app:hyperparameters-preprocessing}
\paragraph{Obtaining Domain Data for \texttt{evolveQA}:}
We scrape domain-specific data using Selenium\footnote{\url{https://selenium-python.readthedocs.io/}}$-$ Python package, from the following websites for building \texttt{evolveQA}. For constructing questions from \textit{Azure} domain, we scraped data from~\url{https://azure.microsoft.com/en-us/updates} website. For \textit{AWS Whatsnew Feed}, we scraped the data from~\url{https://aws.amazon.com/new/}. We note that several links from this website were missing, and as a result, we relied on Wayback Machine~\footnote{\url{https://web.archive.org}} for obtaining the data. Finally, for \textit{WHO-DONs} domain, we scraped the articles from~\url{https://www.who.int/emergencies/disease-outbreak-news}.

\paragraph{Curating \texttt{evolveQA}:}
(1) \textit{Number of concepts $k$:} To determine $k$ during concept extraction, we performed a qualitative study over a randomly sampled set of $10$ documents from each domain. We observed that the documents in cloud domain did not discuss as many different elements as WHO-DONs domain. Therefore, based on the domain and average document length in each domain, we set $k=3$ for \textit{AWS Feed}, and \textit{Azure} domains; we set $k=5$ for \textit{WHO-DONs} domain.

(2) \textit{Additional Pre-processing for WHO-DONs domain:} When processing documents from WHO-DONs domain, we consider only the \textit{virus} entities and all their associated documents for question generation and answer curation. We follow this pre-processing step to ensure that the data for different viruses do not mix with each other. For instance, if we consider a non-virus salient entity such as ``WHO'' and a concept ``Prevention Measures for Disease Outbreaks'', aggregating all documents associated with this entity-concept pair would imply that guidelines for different types of viruses would be aggregated with each other. This does not help with obtaining questions targeting specific piece of knowledge. Therefore, to discard such questions, we focus solely on \textit{virus-specific} entities from this domain.

(3) \textit{Entity Normalization:} Before organizing the corpus based on an entity-concept pair, we also need to ensure that all the entities are in its canonical form. This way, we can potentially reduce errors during the gold answer curation phase, resulting from highly relevant but missing time-sensitive documents. To achieve this, we let the LLM come up with a canonical name during the \textit{concept and salient entity extraction} phase, as described in the prompt in Figure~\ref{prompt-fig:concept-entity-extraction}. Additionally, we also cluster the entities to ensure that all mentions of the same entity are grouped together for the subsequent stages of our pipeline.

(4) \textit{Other Hyperparameters:} We use \texttt{cohere-english}\footnote{\url{https://docs.cohere.com/docs/models}} for embedding all the textual components in our framework. Prior to clustering the textual embeddings with HDBSCAN algorithm, to mitigate the issues with clustering sparse vectors, we reduce their dimensions using UMAP~\cite{mcinnes2018umap} algorithm ($n\_neighbors=15, umap\_components=5$). For HDBSCAN clustering, we use the following hyperparameters: $min\_cluster\_size=5; min\_samples=1; epsilon=0.1$. Finally, for all LLM-calls, we set $temperature=0$.

\paragraph{Evaluating LLMs:} For analyzing the behavior of LLMs on \texttt{evolveQA}, we set $temperature=0, top\_p=0.95$ for all models except \texttt{Llama-3.1-8B-instruct}, and \texttt{GPT-5-mini}. By setting $temperature=0$, we faced token repetition issues with \texttt{Llama-3.1-8B-instruct}, and therefore, for this model, we set $temperature=0.4$ for all our experiments. With \texttt{GPT-5-mini}, the model provider does not allow us to modify temperature parameter and all our experiments with this model is with the default setting (as provided by OpenAI).

\subsection{Membership Inference Attack}
\label{app:mia-subsec}
Table~\ref{tab:mia-attack-evidence} shows the results from MIA using \textit{min-k\%} algorithm. We conducted this experiment with \texttt{Llama-3.1-8B-Instruct} for $k=0.2$. First, for each domain, we randomly sampled $500$ data points ($500$ from \textit{before} the model cut-off period, and $500$ from after the model cut-off period). We applied \textit{min-k\%} algorithm on both these data samples and observed a significant gap in the detection rates across these two sets, indicating that models do not possess knowledge beyond their cut-off period.
\begin{table}[ht!]
\centering
\resizebox{\columnwidth}{!}{%
\begin{tabular}{l|c|c}
\toprule
\multirow{2}{*}{\textbf{Domain}} & \multicolumn{2}{c}{\textbf{Min-K\% Prob}} \\
\cmidrule(lr){2-3}
& \textbf{\% Members (Before Cutoff)} & \textbf{\% Non-Members (After Cutoff)} \\
\midrule
Azure & 81 & 70.6 \\
AWS & 76.94 & 50.38 \\
WHO & 81.8 & 63.29 \\
\bottomrule
\end{tabular}%
}
\caption{\small Validation of pre-training exposure via Membership Inference Attack (MIA). Using the \texttt{min-k\%} algorithm ($k=0.2$), \texttt{Llama-3.1-8B-Instruct} was tasked to differentiate between 500 randomly sampled member (pre-cutoff) and 500 non-member (post-cutoff) documents. The resulting $11\%$ to $26\%$ point gap in detection rates between members and non-members across all domains confirms that our benchmark targets knowledge represented within the model's parameters.}
\label{tab:mia-attack-evidence}
\end{table}

\subsection{Human Evaluation}
\label{subsec:app-human-eval}
This section discusses results from our human evaluation: (1) human assessment on the quality of the benchmark (\texttt{evolveQA}); (2) meta evaluation of our model judge (\texttt{Claude Sonnet 4}).
\subsubsection{Evaluating \texttt{evolveQA}: Evidence from Human Assessment}
\label{app-subsubsec:evolveQA-human-eval}
To assess the quality of \texttt{evolveQA}, we conducted a human evaluation. For this evaluation, we sampled $100$ question-answer pairs considering the validation of the LLM-judge. We used the same sample to perform a meta evaluation of the model judge (Section~\ref{app-subsec:model-judge-meta-human-eval}). In particular, we randomly sampled $50$ instances from the cases where the LLM-judge provided a \textit{CORRECT\_AND\_CURRENT} label as the judgement, $30$ randomly sampled from the cases where the LLM-judge provided \textit{INCORRECT} as the judgement, $20$ randomly sampled from the cases where the LLM-judge provided \textit{OUTDATED} label as the judgement; across different domains and different model cutoffs. Four annotators (among the authors of this paper) were tasked with evaluating two key components: the generated questions and the curated gold answers.

For \textbf{Question Evaluation}, we assessed three dimensions: (a) \textbf{Groundedness:} measures if the question is faithfully grounded in its corresponding {entity, concept, attribute} (E-C-A) tuple. (b) \textbf{Knowledge Evolution:} validates if the attribute targeted by the question genuinely evolves over time, ensuring the question is relevant to our study of temporal knowledge conflicts. (c) \textbf{Quality \& Specificity:} evaluates if the question targets non-trivial, specific information, avoiding generic or overly simple queries.

For \textbf{Gold Answer Correctness}, annotators verified the factual correctness of the curated answers by cross-referencing them against both the provided top-$7$ documents from source corpora and external search engine results.

Each datapoint was annotated by one annotator. Table~\ref{tab:human-evaluation-results} shows the results from human evaluation. It confirms the high quality of our generation pipeline. Specifically, the questions from our benchmark demonstrate high performance on individual quality dimensions: $92\%$ in groundedness, and $94\%$ for attributes that genuinely evolve over time, validating the effectiveness of our temporal conflict identification process. Further, in \textbf{Overall Correctness}$-$representing the proportion of questions that pass \textit{all} three question quality dimensions simultaneously, achieves a high score of $80\%$. This high question quality score is complemented by $91.57\%$ correctness rate for the curated gold answers. Taken together, these results demonstrate that \texttt{evolveQA} is a reliable benchmark.

\subsubsection{Meta Evaluation of Model Judge} 
\label{app-subsec:model-judge-meta-human-eval}
In a supplementary analysis, on this sample of $100$ question-answer pairs, we also evaluated the performance of our model judge (\texttt{Claude Sonnet 4}). The judge achieved an accuracy of $95.9\%$ in evaluating LLM responses against the gold answers, confirming its reliability for our experiments. We also found that in $2.7\%$ of the cases, the judge failed to correctly identify whether the model's response was ``outdated'' or ``incorrect''. In the remaining $1.35\%$ of the cases, the judge was completely incorrect in its decisions.

\begin{table}[ht!]
\centering
\begin{tabular}{c c}
\toprule
\textbf{Metric} & \textbf{Score (\%)} \\
\midrule
\multicolumn{1}{c}{\textbf{Question Evaluation}} \\
\midrule
\quad Grounded in E-C-A & 92 \\
\quad Knowledge Evolution & 94 \\
\quad Quality \& Specificity & 89 \\
\quad \textbf{Overall Correctness} & 80 \\
\midrule
\textbf{Gold Answer Correctness} & 91.57 \\
\bottomrule
\end{tabular}
\caption{ \small
Human evaluation results validating the quality of the \texttt{evolveQA} benchmark. We observe high quality on individual dimensions, such as \textit{grounding} (92\%) and targeting genuine \textit{knowledge evolution} (94\%). \textbf{Overall Correctness} is our strict composite metric, representing the proportion of QA pairs that successfully pass \textit{all} preceding quality checks. The high score on this strict measure (80\%), coupled with high gold answer correctness (91.57\%), confirms the overall reliability of our framework.
}
\label{tab:human-evaluation-results}
\end{table}

\subsection{Main Results: Additional Discussion}
\begin{figure}[t!]
\includegraphics[width=\linewidth]{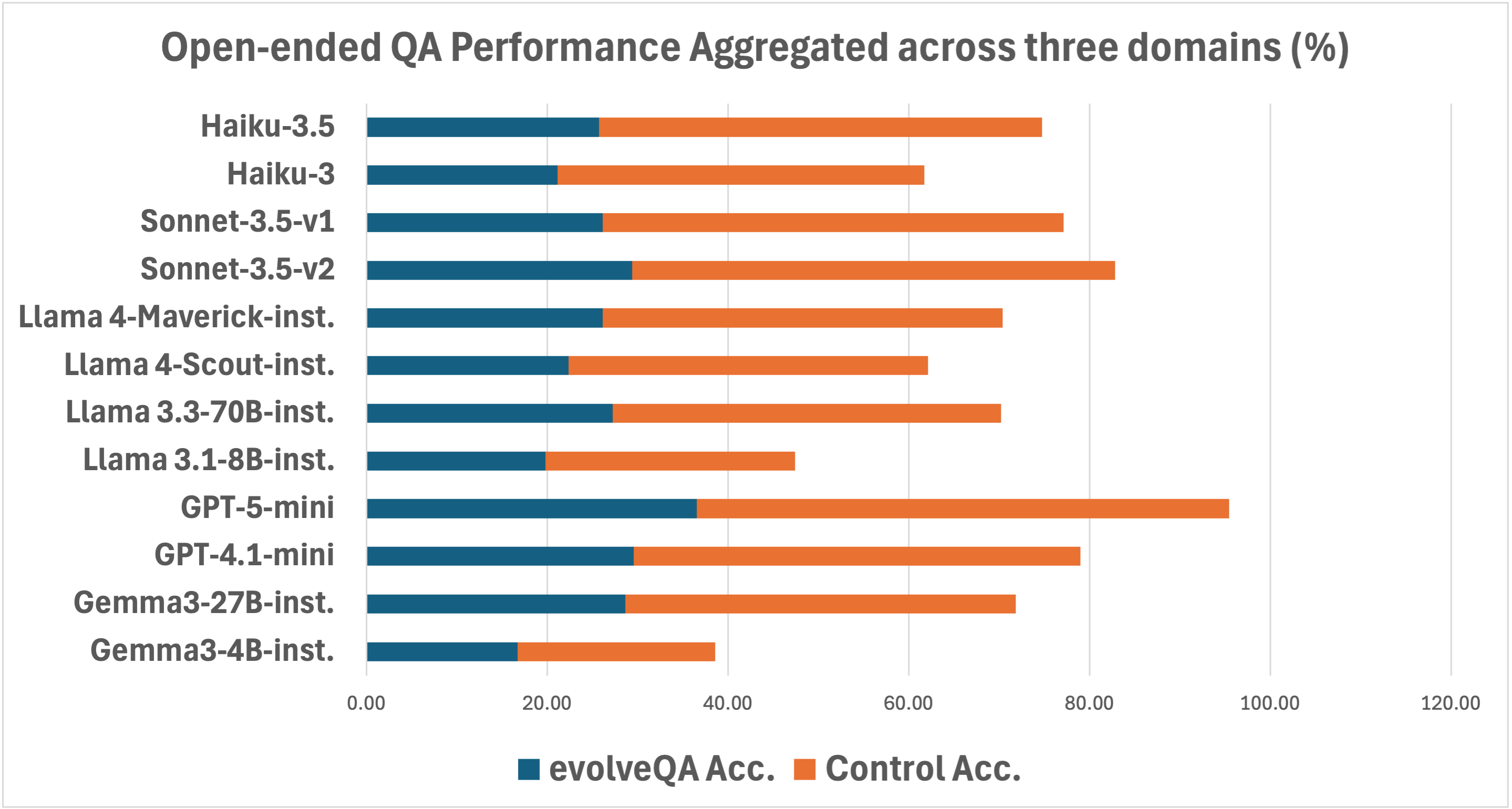}
\small \caption{Performance comparison between \texttt{evolveQA} and control set on open-ended QA, aggregated across the three domains. We observe a significant performance degredation on \texttt{evolveQA} across all models.}
\label{fig:performance-degradation-aggregate-evolveQA}
\end{figure}

Figure~\ref{fig:performance-degradation-aggregate-evolveQA} illustrates the performance of \texttt{evolveQA} when compared against the control set. For the purposes of this visualization, we have averaged model performance across all the domains. We observe that each model significantly lags behind the control set when probed for temporally conflicting information in an open-ended QA setting.

\subsubsection{Notable Anomalies: \texttt{Claude-3.5-Haiku}}
\label{app-subsubsec:notable-anomalies}
\begin{figure*}[t!]
\centering
\includegraphics[width=\linewidth]{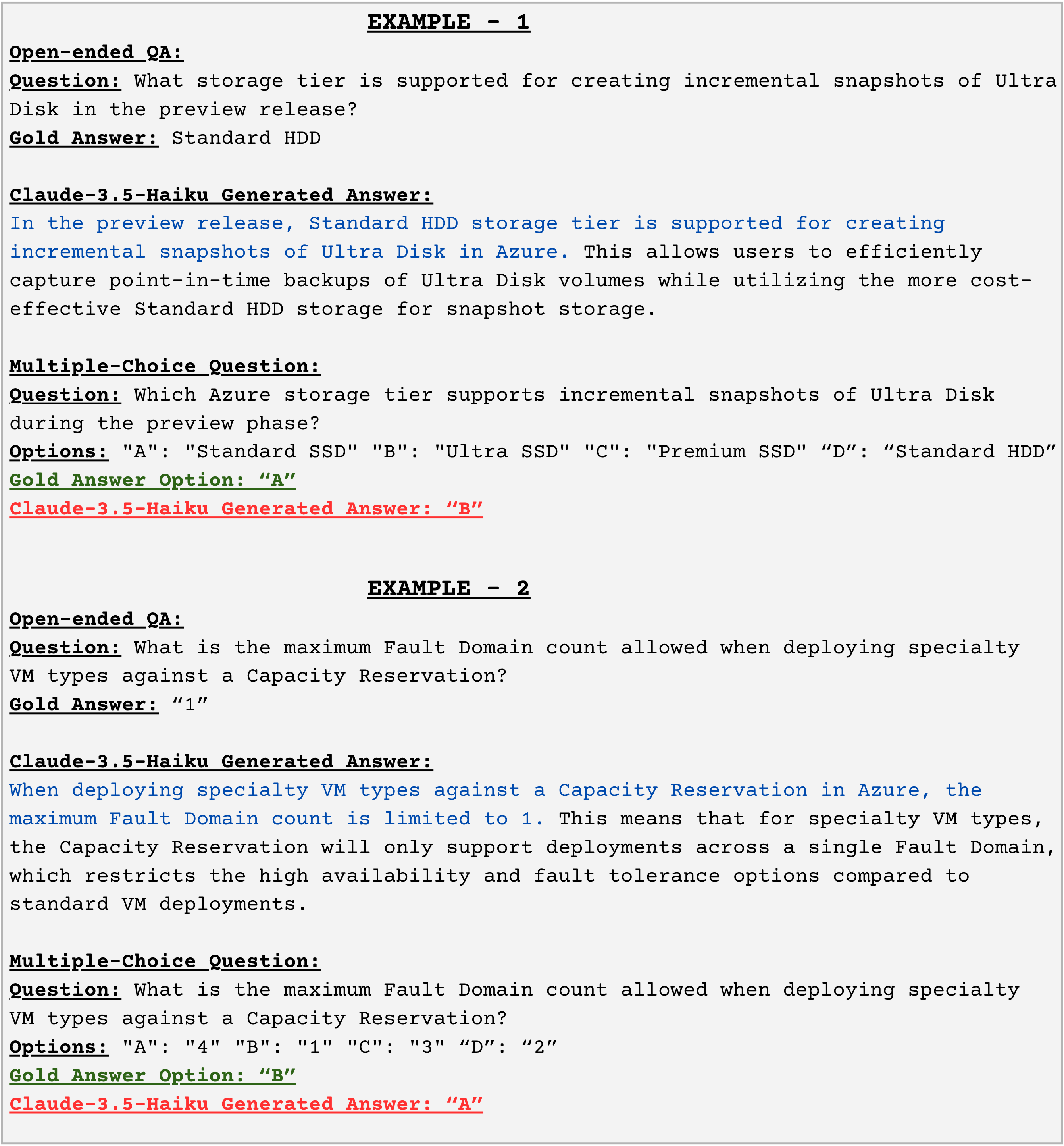}
\small \caption{Shows two examples where \texttt{Claude-3.5-Haiku} generates correct response in open-ended QA whereas, it generates an incorrect response for its corresponding MCQ format in the control set.}
\label{app-fig:performance-discrepancy-claude-3.5-haiku}
\end{figure*}

Here, we provide examples of our qualitative study. Figure~\ref{app-fig:performance-discrepancy-claude-3.5-haiku} illustrates two cases from \textit{Azure} domain, where the model (\texttt{Claude-3.5-Haiku}) generated correct response in open-ended QA and generated incorrect response in its corresponding MCQ format.

\section{Ablation Study Results}
In this section, we provide the detailed results for the following ablation studies: (1) Outdated Response Rate (Section~\ref{app-subsec:orr}); (2) Performance on \textit{superseding} knowledge (Section~\ref{app-subsec:superseding-knowledge}); (3) Performance with explicit context (Section~\ref{app-subsec:eca-context}).
In addition, to mitigate self-preference bias, we also evaluate responses from Claude models with a different model-judge$-$~\texttt{GPT-4.1-mini} (Section~\ref{app-subsec:diff-model-judge}).
\subsection{Outdated Response Rate}
\label{app-subsec:orr}
As discussed in the ablation study in the main paper (Section~\ref{subsec:results-and-ablations}), simply measuring accuracy on the open-ended QA task can mask the specific failure modes of LLMs on \texttt{evolveQA}. To provide a more granular analysis, we categorize each generated response into one of three categories:
(a) \textbf{Accurate:} The response is factually correct and aligns with the most current gold-standard answer.
(b) \textbf{Outdated:} The response is factually correct for a past time period but is now outdated or has been superseded. This is the Outdated Response Rate (\texttt{ORR}).
(c) \textbf{Incorrect:} The response is neither accurate nor outdated.
Table~\ref{tab:evolveqa-breakdown-orr} presents the full breakdown for all models on the open-ended QA task. The key observation is the consistently high \texttt{ORR}, which confirms that a primary failure mode is the retrieval of stale, parametric knowledge. We observe that for certain models, the proportion of outdated responses reaches as high as $55\%$ in the \textit{AWS} domain, $50\%$ in the \textit{Azure} domain, and $39\%$ in the \textit{WHO-DONs} domain. This data provides strong evidence that LLMs frequently mistake deeply ingrained yet obsolete information for current fact.
\begin{table*}[!htb]
\small 
\centering
\resizebox{1.95\columnwidth}{!}{%
\begin{tabular}
{>{\centering\arraybackslash}m{2.5cm}|>{\arraybackslash}m{2.5cm}|>{\arraybackslash}m{4.5cm}|>{\centering\arraybackslash}m{1.8cm}|>{\centering\arraybackslash}m{2.0cm}|>{\centering\arraybackslash}m{1.8cm}||>{\centering\arraybackslash}m{2.0cm}}
\toprule
& & & \multicolumn{3}{c||}{\textbf{\texttt{evolveQA} (\%)}} & \multicolumn{1}{c}{\textbf{Control Set (\%)}} \\ 
\cmidrule{4-6} \cmidrule{7-7}
\textbf{Domain} & \textbf{Model Family} & \textbf{Model} & \textbf{Acc.} & \textbf{Outdated (\texttt{ORR})} & \textbf{Incorrect} & \textbf{Acc.} \\
\midrule

\multirow{12}{*}{\textbf{\rotatebox[origin=c]{90}{AWS Whatsnew Feed}}}    
& \multirow{4}{*}{Anthropic} 
& Claude 3.5-sonnet-v2 & 26.95 & 50.21 & 22.84 & 46.39 \\
& & Claude 3.5-sonnet-v1 & 22.9 & 51.33 & 25.77 & 44.41 \\
& & Claude 3 Haiku & 15.93 & 52.46 & 31.62 & 33.84 \\
& & Claude 3.5 Haiku & 22.64 & 54.91 & 22.45 & 42.14 \\
\cmidrule{2-7}
& \multirow{4}{*}{Llama} 
& Llama 3.1-8B-Instruct & 14.44 & 42.89 & 42.67 & 20.33 \\
& & Llama 3.3-70B-Instruct & 18.53 & 47.63 & 33.84 & 33.13 \\
& & Llama 4-Scout-instruct & 17.09 & 51.45 & 31.45 & 31.78 \\
& & Llama 4-Maverick-instruct & 22.18 & 47.64 & 30.18 & 36.45 \\
\cmidrule{2-7}
& \multirow{2}{*}{OpenAI} 
& GPT-4.1-mini & 22.86 & 54.67 & 22.47 & 41.69 \\
& & GPT-5-mini & 29.62 & 42.94 & 27.44 & 55.4 \\
\cmidrule{2-7}
& \multirow{2}{*}{Gemma} 
& Gemma3-4B-instruct & 12.18 & 36.73 & 51.09 & 18.83 \\
& & Gemma3-27B-instruct & 21.82 & 46.18 & 32.0 & 34.71 \\
\midrule

\multirow{12}{*}{\textbf{\rotatebox[origin=c]{90}{Azure Updates}}}    
& \multirow{4}{*}{Anthropic} 
& Claude 3.5-sonnet-v2 & 22.95 & 50.27 & 26.78 & 49.82 \\
& & Claude 3.5-sonnet-v1 & 21.31 & 46.99 & 31.69 & 46.24 \\
& & Claude 3 Haiku & 18.18 & 46.75 & 35.06 & 27.1 \\
& & Claude 3.5 Haiku & 21.32 & 48.22 & 30.46 & 41.5 \\
\cmidrule{2-7}
& \multirow{4}{*}{Llama} 
& Llama 3.1-8B-Instruct & 17.58 & 34.55 & 47.88 & 16.85 \\
& & Llama 3.3-70B-Instruct & 24.24 & 44.24 & 31.52 & 36.26 \\
& & Llama 4-Scout-instruct & 18.72 & 48.82 & 33.0 & 31.58 \\
& & Llama 4-Maverick-instruct & 23.15 & 46.31 & 30.54 & 37.19 \\
\cmidrule{2-7}
& \multirow{2}{*}{OpenAI} 
& GPT-4.1-mini & 26.84 & 40.53 & 32.63 & 38.35 \\
& & GPT-5-mini & 29.47 & 45.26 & 25.26 & 49.82 \\
\cmidrule{2-7}
& \multirow{2}{*}{Gemma} 
& Gemma3-4B-instruct & 14.29 & 30.54 & 55.17 & 15.09 \\
& & Gemma3-27B-instruct & 25.12 & 39.41 & 35.47 & 30.88 \\
\midrule

\multirow{12}{*}{\textbf{\rotatebox[origin=c]{90}{WHO-Disease Outbreak News}}}    
& \multirow{4}{*}{Anthropic} 
& Claude 3.5-sonnet-v2 & 38.41 & 39.07 & 22.52 & 64.09 \\
& & Claude 3.5-sonnet-v1 & 34.23 & 38.93 & 26.85 & 62.43 \\
& & Claude 3 Haiku & 29.41 & 38.97 & 31.62 & 60.69 \\
& & Claude 3.5 Haiku & 33.33 & 37.74 & 28.93 & 63.44 \\
\cmidrule{2-7}
& \multirow{4}{*}{Llama} 
& Llama 3.1-8B-Instruct & 27.4 & 38.36 & 34.25 & 45.71 \\
& & Llama 3.3-70B-Instruct & 39.04 & 37.67 & 23.29 & 59.43 \\
& & Llama 4-Scout-instruct & 31.36 & 44.97 & 23.67 & 55.91 \\
& & Llama 4-Maverick-instruct & 33.14 & 37.87 & 28.99 & 59.14 \\
\cmidrule{2-7}
& \multirow{2}{*}{OpenAI} 
& GPT-4.1-mini & 39.1 & 35.26 & 25.64 & 68.13 \\
& & GPT-5-mini & 50.64 & 30.77 & 18.59 & 71.43 \\
\cmidrule{2-7}
& \multirow{2}{*}{Gemma} 
& Gemma3-4B-instruct & 23.67 & 23.08 & 53.25 & 31.72 \\
& & Gemma3-27B-instruct & 39.05 & 31.95 & 28.99 & 63.98 \\
\bottomrule

\end{tabular}}
\caption{\small Breakdown of performance on \texttt{evolveQA} Open-Ended QA. Here, we report: Accuracy, \textit{Outdated Response Rate} (\texttt{ORR}), and Incorrect response (\%), across datasets. Performance on non-conflicting control set is also reported. We observe a high \texttt{ORR}, at least $23\%$, across all datasets.}
\label{tab:evolveqa-breakdown-orr}
\end{table*}

\subsection{Superseding Knowledge}
\label{app-subsec:superseding-knowledge}
Table~\ref{tab:knowledge_types_examples} provides an example of \textit{additive} vs. \textit{superseding} knowledge. In the case of \textit{additive} knowledge, there could potentially be multiple correct answers. As a result, model's have a higher chance of getting them correct. However, \textit{superseding} knowledge becomes highly challenging for the models, especially in MCQ format, where we present a previously valid response as part of the input question.

Consequently, as illustrated in Table~\ref{tab:evolveQA-superseding-ablation}, we observe a significant reduction in performance in the case of \textit{superseding} knowledge, when compared against the entire \texttt{evolveQA} benchmark.

\begin{table*}[ht!]
\centering
\resizebox{\textwidth}{!}{%
\begin{tabular}{p{6.5cm} l|p{6.5cm} l}
\toprule
\multicolumn{2}{c|}{\textbf{Additive Knowledge}} & \multicolumn{2}{c}{\textbf{Superseding Knowledge}} \\
\midrule
\textbf{Question} & \textbf{Answers (all valid)} & \textbf{Question} & \textbf{Answer} \\
\midrule
\multirow{3}{6.5cm}{How does Amazon Kendra obtain user and group access control information for search filtering?} & through secure search tokens... & \multirow{3}{6.5cm}{How many AWS regions globally support Amazon Kendra?} & 4 \\
& through its principal store... & & 5 \\
& AWS SSO identity store... & & 7 \\
\bottomrule
\end{tabular}%
}
\caption{\small This table provides examples of questions that test for two distinct types of knowledge: \textit{Additive}$-$multiple valid answers, and \textit{Superseding}$-$ only most recent answer is valid.}
\label{tab:knowledge_types_examples}
\end{table*}

\begin{table*}[ht!]
\small 
\centering
\resizebox{2.1\columnwidth}{!}{%
\begin{tabular}
{>{\centering\arraybackslash}m{2.5cm}|>{\arraybackslash}m{2.5cm}|>{\arraybackslash}m{3.8cm}|
>{\centering\arraybackslash}m{2.5cm}>{\centering\arraybackslash}m{1.8cm}>{\centering\arraybackslash}m{2.3cm}|
>{\centering\arraybackslash}m{2.5cm}>{\centering\arraybackslash}m{1.8cm}>{\centering\arraybackslash}m{2.3cm}} 
\toprule
& & & \multicolumn{3}{c|}{\textbf{MCQ: Acc. (\%)}} & \multicolumn{3}{c}{\textbf{Open-Ended QA: Acc. (\%)}} \\
\cmidrule{4-9}
\textbf{Domain} & \textbf{Model Family} & \textbf{Model} & \textbf{\texttt{evolveQA - superseding} knowledge} & \textbf{\texttt{evolveQA}} & \textbf{Control Set} & \textbf{\texttt{evolveQA - superseding} knowledge} & \textbf{\texttt{evolveQA}} & \textbf{Control Set} \\
\midrule

\multirow{12}{*}{\textbf{\rotatebox[origin=c]{90}{AWS Whatsnew Feed}}}
& \multirow{4}{*}{Anthropic}
& Claude 3.5-sonnet-v2 & 51.32 & 63.33 & 82.49 & 25.28 & 26.95 & 46.39 \\
& & Claude 3.5-sonnet-v1 & 49.81 & 63.33 & 83.47 & 19.78 & 22.9 & 44.41 \\
& & Claude 3 Haiku & 49.79 & 62.47 & 68.07 & 15.88 & 15.93 & 33.84 \\
& & Claude 3.5 Haiku & 60.93 & 67.69 & 73.24 & 21.51 & 22.64 & 42.14 \\
\cmidrule{2-9}
& \multirow{4}{*}{Llama}
& Llama 3.1-8B-Instruct & 53.7 & 61.84 & 67.53 & 15.56 & 14.44 & 20.33 \\
& & Llama 3.3-70B-Inst. & 56.42 & 64.99 & 73.78 & 15.95 & 18.53 & 33.13 \\
& & Llama 4-Scout-inst. & 58.54 & 68.32 & 75.35 & 16.72 & 17.09 & 31.78 \\
& & Llama 4-Maverick-inst. & 53.66 & 66.3 & 79.51 & 21.95 & 22.18 & 36.45 \\
\cmidrule{2-9}
& \multirow{2}{*}{OpenAI}
& GPT-4.1-mini & 47.76 & 62.78 & 78.24 & 17.54 & 22.86 & 41.69 \\
& & GPT-5-mini & 47.01 & 59.86 & 82.08 & 20.52 & 29.62 & 55.4 \\
\cmidrule{2-9}
& \multirow{2}{*}{Gemma}
& Gemma3-4B-inst. & 42.16 & 52.85 & 58.24 & 11.85 & 12.12 & 18.83 \\
& & Gemma3-27B-inst. & 52.26 & 64.46 & 71.65 & 19.86 & 21.82 & 34.71 \\
\midrule

\multirow{12}{*}{\textbf{\rotatebox[origin=c]{90}{Azure Updates}}}
& \multirow{4}{*}{Anthropic}
& Claude 3.5-sonnet-v2 & 64.44 & 70.33 & 77.73 & 23.33 & 22.95 & 49.82 \\
& & Claude 3.5-sonnet-v1 & 63.33 & 70.88 & 78.91 & 18.89 & 21.31 & 46.24 \\
& & Claude 3 Haiku & 66.27 & 70.78 & 64.44 & 13.25 & 18.18 & 27.1 \\
& & Claude 3.5 Haiku & 56.84 & 68.72 & 64.75 & 15.79 & 21.32 & 41.5 \\
\cmidrule{2-9}
& \multirow{4}{*}{Llama}
& Llama 3.1-8B-Inst. & 56.18 & 63.64 & 60.4 & 19.1 & 17.58 & 16.85 \\
& & Llama 3.3-70B-Inst. & 59.55 & 70.3 & 73.2 & 21.35 & 24.24 & 36.26 \\
& & Llama4-Scout-inst. & 61.86 & 72.64 & 71.76 & 19.59 & 18.72 & 31.58 \\
& & Llama4-Maverick-inst. & 68.04 & 76.12 & 75.95 & 24.74 & 23.15 & 37.19 \\
\cmidrule{2-9}
& \multirow{2}{*}{OpenAI}
& GPT-4.1-mini & 59.14 & 71.81 & 76.56 & 22.58 & 26.84 & 38.35 \\
& & GPT-5-mini & 52.69 & 67.55 & 76.95 & 26.88 & 29.47 & 49.82 \\
\cmidrule{2-9}
& \multirow{2}{*}{Gemma}
& Gemma3-4B-inst. & 50.52 & 54.73 & 51.53 & 14.43 & 14.29 & 15.09 \\
& & Gemma3-27B-inst. & 62.89 & 69.15 & 64.12 & 22.68 & 25.12 & 30.88 \\
\midrule

\multirow{10}{*}{\textbf{\rotatebox[origin=c]{90}{WHO–Disease Outbreak News}}}
& \multirow{4}{*}{Anthropic}
& Claude 3.5-sonnet-v2 & 64.71 & 73.1 & 86.19 & 31.03 & 38.41 & 64.09 \\
& & Claude 3.5-sonnet-v1 & 62.35 & 71.72 & 90.61 & 25.88 & 34.23 & 62.43 \\
& & Claude 3 Haiku & 62.03 & 66.67 & 81.5 & 24.05 & 29.41 & 60.69 \\
& & Claude 3.5 Haiku & 70.1 & 72.08 & 80.11 & 28.87 & 33.33 & 63.44 \\
\cmidrule{2-9}
& \multirow{4}{*}{Llama}
& Llama 3.1-8B-Inst. & 62.79 & 67.36 & 73.14 & 23.26 & 27.4 & 45.71 \\
& & Llama 3.3-70B-Inst. & 61.63 & 68.75 & 83.43 & 32.56 & 39.04 & 59.43 \\
& & Llama 4-Scout-inst. & 62.11 & 68.71 & 82.26 & 29.47 & 31.36 & 55.91 \\
& & Llama 4-Maverick-inst. & 70.53 & 76.07 & 84.95 & 28.42 & 33.14 & 59.14 \\
\cmidrule{2-9}
& \multirow{2}{*}{OpenAI}
& GPT-4.1-mini & 57.3 & 72.19 & 81.87 & 33.71 & 39.1 & 68.13 \\
& & GPT-5-mini & 59.55 & 72.85 & 86.26 & 42.7 & 50.64 & 71.43 \\
\cmidrule{2-9}
& \multirow{2}{*}{Gemma}
& Gemma3-4B-inst. & 56.84 & 63.19 & 64.52 & 18.95 & 23.67 & 31.72 \\
& & Gemma3-27B-inst. & 67.37 & 71.17 & 79.51 & 34.74 & 39.05 & 63.98 \\
\bottomrule
\end{tabular}}
\caption{\small Performance comparison on the \texttt{superseding} knowledge subset of \texttt{evolveQA}, the full \texttt{evolveQA} benchmark, and the non-conflicting control dataset. The sharp decline in MCQ accuracy on the superseding subset highlights that historically correct but outdated answers significantly impairs the models' ability to identify current information.}
\label{tab:evolveQA-superseding-ablation}
\end{table*}

\subsection{Providing Explicit \textit{Entity, Concept, Attribute} Context}
\label{app-subsec:eca-context}
Table~\ref{tab:evolveQA-context-ablation} shows the results from the cases where we provide explicit contextual$-$\{\textit{entity, concept, attribute}\} and temporal hints (through a ``current date'' $-$ which is a randomly chosen date before the model's knowledge cut-off date and based on the question's time-frame).
\begin{table*}[ht!]
\small 
\centering
\resizebox{\textwidth}{!}{%
\begin{tabular}{>{\centering\arraybackslash}m{1.8cm}|c|l|c|c|c|c}
\toprule
\textbf{Domain} & \textbf{Model Family} & \textbf{Model} & \multicolumn{4}{c}{\textbf{Accuracy (\%)}} \\
\cmidrule{4-7}
& & & \textbf{Open-ended QA} & \textbf{Open-Ended QA with Context} & \textbf{Open-Ended QA with Context and Date} & \textbf{Open-Ended QA with Date} \\
\midrule
\multirow{12}{*}{\textbf{\rotatebox[origin=c]{90}{AWS Whatsnew Feed}}} & \multirow{4}{*}{\textbf{Anthropic}} & Claude 3.5-sonnet-v2 & 26.95 & 31.69 & 32.3 & 28.4 \\
& & Claude 3.5-sonnet-v1 & 22.9 & 27.37 & 28.83 & 26.34 \\
& & Claude 3 Haiku & 15.93 & 21.78 & 24.12 & 16.16 \\
& & Claude 3.5 Haiku & 22.64 & 25.66 & 28.49 & 26.23 \\
\cmidrule{2-7}
& \multirow{4}{*}{\textbf{Llama}} & Llama 3.1-8B-Instruct & 14.44 & 20.47 & 20.52 & 15.55 \\
& & Llama 3.3-70B-Instruct & 18.53 & 23.28 & 26.08 & 20.26 \\
& & Llama 4-Scout-instruct & 17.09 & 23.82 & 24.36 & 19.27 \\
& & Llama 4-Maverick-instruct & 22.18 & 24.91 & 26.36 & 22.18 \\
\cmidrule{2-7}
& \multirow{2}{*}{\textbf{OpenAI}} & GPT-4.1-mini & 22.86 & 25.84 & 28.83 & 24.25 \\
& & GPT-5-mini & 29.62 & 34 & 34.19 & 29.82 \\
\cmidrule{2-7}
& \multirow{2}{*}{\textbf{Gemma}} & Gemma3-4B-instruct & 12.18 & 18.55 & 20.91 & 14.36 \\
& & Gemma3-27B-instruct & 21.82 & 27.09 & 27.09 & 23.82 \\
\midrule
\multirow{12}{*}{\textbf{\rotatebox[origin=c]{90}{Azure Updates}}} & \multirow{4}{*}{\textbf{Anthropic}} & Claude 3.5-sonnet-v2 & 22.95 & 31.15 & 31.15 & 22.4 \\
& & Claude 3.5-sonnet-v1 & 21.31 & 32.24 & 30.05 & 22.95 \\
& & Claude 3 Haiku & 18.18 & 27.27 & 25.97 & 21.43 \\
& & Claude 3.5 Haiku & 21.32 & 30.96 & 30.96 & 23.35 \\
\cmidrule{2-7}
& \multirow{4}{*}{\textbf{Llama}} & Llama 3.1-8B-Instruct & 17.58 & 24.24 & 26.67 & 16.36 \\
& & Llama 3.3-70B-Instruct & 24.24 & 28.48 & 29.7 & 25.45 \\
& & Llama 4-Scout-instruct & 18.72 & 27.59 & 27.59 & 19.7 \\
& & Llama 4-Maverick-instruct & 23.15 & 32.51 & 31.53 & 23.65 \\
\cmidrule{2-7}
& \multirow{2}{*}{\textbf{OpenAI}} & GPT-4.1-mini & 26.84 & 31.58 & 31.58 & 29.47 \\
& & GPT-5-mini & 29.47 & 36.32 & 36.84 & 28.95 \\
\cmidrule{2-7}
& \multirow{2}{*}{\textbf{Gemma}} & Gemma3-4B-instruct & 14.29 & 21.18 & 20.2 & 15.27 \\
& & Gemma3-27B-instruct & 25.12 & 29.06 & 27.09 & 25.12 \\
\midrule
\multirow{12}{*}{\textbf{\rotatebox[origin=c]{90}{WHO-Disease Outbreak News}}} & \multirow{4}{*}{\textbf{Anthropic}} & Claude 3.5-sonnet-v2 & 38.41 & 39.6 & 46.31 & 39.6 \\
& & Claude 3.5-sonnet-v1 & 34.23 & 39.6 & 38.93 & 36.91 \\
& & Claude 3 Haiku & 29.41 & 30.15 & 33.82 & 31.62 \\
& & Claude 3.5 Haiku & 33.33 & 34.59 & 32.7 & 36.48 \\
\cmidrule{2-7}
& \multirow{4}{*}{\textbf{Llama}} & Llama 3.1-8B-Instruct & 27.4 & 28.08 & 26.03 & 27.4 \\
& & Llama 3.3-70B-Instruct & 39.04 & 36.99 & 39.04 & 34.93 \\
& & Llama 4-Scout-instruct & 31.36 & 29.59 & 30.77 & 29.59 \\
& & Llama 4-Maverick-instruct & 33.14 & 33.14 & 33.14 & 31.95 \\
\cmidrule{2-7}
& \multirow{2}{*}{\textbf{OpenAI}} & GPT-4.1-mini & 39.1 & 39.74 & 41.03 & 37.18 \\
& & GPT-5-mini & 50.64 & 49.36 & 53.21 & 46.15 \\
\cmidrule{2-7}
& \multirow{2}{*}{\textbf{Gemma}} & Gemma3-4B-instruct & 23.67 & 26.04 & 22.49 & 28.4 \\
& & Gemma3-27B-instruct & 39.05 & 34.32 & 37.28 & 40.24 \\
\bottomrule
\end{tabular}%
}
\caption{\small Ablation study on the impact of providing contextual and temporal hints to LLMs in the open-ended QA setting of \texttt{evolveQA}. The table presents the accuracy of each model across the three domains under four conditions, comparing the baseline (no context) to performance when provided with context (the \{\textit{entity, concept, attribute}\} tuple), a date hint, or both. The date hint is sampled relative to the model's knowledge cut-off date.}
\label{tab:evolveQA-context-ablation}
\end{table*}

\subsection{Mitigating Self-Preference Bias: Evaluating Claude Responses with \texttt{GPT-4.1-mini}}
\label{app-subsec:diff-model-judge}
To evaluate open-ended responses, our methodology uses \texttt{Claude Sonnet 4} (from \textit{Anthropic family}) as the model judge. We utilize this model-judge even to evaluate responses from other \textit{Anthropic family} of models such as \texttt{Claude 3.5 Sonnet, Haiku}. Therefore, to empirically validate if this model-judge has a tendency to prefer responses from its own family, we evaluate open-ended responses from \textit{Anthropic family} of models using an additional model-judge$-$\texttt{GPT-4.1-mini}. Even with this judge, we observe a consistent pattern (illustrated in Table~\ref{tab:evolveQA-judge-eval-ablation}), as discussed in the main paper.
\begin{table*}[ht!]
\small 
\centering
\resizebox{\textwidth}{!}{%
\begin{tabular}{c|c|l||cc||ccc|ccc}
\toprule
\multicolumn{3}{c||}{} & \multicolumn{2}{c||}{\textbf{Non-Conflicting Control Dataset}} & \multicolumn{6}{c}{\textbf{\texttt{evolveQA} [Open-Ended QA (\%)]}} \\
\multicolumn{3}{c||}{} & \multicolumn{2}{c||}{\textbf{[Open-Ended QA (\%)]}} & \multicolumn{6}{c}{} \\
\cmidrule(lr){4-5} \cmidrule(lr){6-11}
\textbf{Domain} & \textbf{Model Family} & \textbf{Model} & \textbf{Sonnet-4 Judge} & \textbf{GPT-4.1-mini Judge} & \multicolumn{3}{c|}{\textbf{Sonnet-4 Judge}} & \multicolumn{3}{c}{\textbf{GPT-4.1-mini Judge}} \\
\cmidrule(lr){4-5} \cmidrule(lr){6-8} \cmidrule(lr){9-11}
& & & \textbf{Acc.} & \textbf{Acc.} & \textbf{Acc.} & \textbf{Outdated} & \textbf{Incorrect} & \textbf{Acc.} & \textbf{Outdated} & \textbf{Incorrect} \\
\midrule
\multirow{4}{*}{\textbf{\rotatebox[origin=c]{90}{Azure}}} & \multirow{4}{*}{Anthropic} & Claude 3.5-sonnet-v2 & 49.82 & 54.12 & 22.95 & 50.27 & 26.78 & 28.89 & 47.78 & 23.33 \\
& & Claude 3.5-sonnet-v1 & 46.24 & 51.61 & 21.31 & 46.99 & 31.69 & 20.00 & 56.67 & 23.33 \\
& & Claude 3 Haiku & 27.10 & 33.21 & 18.18 & 46.75 & 35.06 & 25.30 & 53.01 & 21.69 \\
& & Claude 3.5 Haiku & 41.50 & 43.66 & 21.32 & 48.22 & 30.46 & 25.26 & 51.58 & 23.16 \\
\midrule
\multirow{4}{*}{\textbf{\rotatebox[origin=c]{90}{AWS}}} & \multirow{4}{*}{Anthropic} & Claude 3.5-sonnet-v2 & 46.39 & 50.92 & 26.95 & 50.21 & 22.84 & 29.43 & 59.62 & 10.94 \\
& & Claude 3.5-sonnet-v1 & 44.41 & 49.65 & 22.90 & 51.33 & 25.77 & 25.94 & 60.90 & 13.16 \\
& & Claude 3 Haiku & 33.84 & 37.69 & 15.93 & 52.46 & 31.62 & 21.94 & 62.87 & 15.19 \\
& & Claude 3.5 Haiku & 42.14 & 46.88 & 22.64 & 54.91 & 22.45 & 26.79 & 59.29 & 13.93 \\
\midrule
\multirow{4}{*}{\textbf{\rotatebox[origin=c]{90}{WHO}}} & \multirow{4}{*}{Anthropic} & Claude 3.5-sonnet-v2 & 64.09 & 69.61 & 38.41 & 39.07 & 22.52 & 40.00 & 41.18 & 18.82 \\
& & Claude 3.5-sonnet-v1 & 62.43 & 64.64 & 34.23 & 38.93 & 26.85 & 42.35 & 40.00 & 17.65 \\
& & Claude 3 Haiku & 60.69 & 56.65 & 29.41 & 38.97 & 31.62 & 29.11 & 41.77 & 29.11 \\
& & Claude 3.5 Haiku & 63.44 & 63.44 & 33.33 & 37.74 & 28.93 & 37.11 & 21.65 & 41.24 \\
\bottomrule
\end{tabular}%
}
\caption{\small Shows accuracy scores for the Claude-family models on open-ended \texttt{evolveQA}, evaluated by our primary judge (\texttt{Claude Sonnet 4}) and cross-validated with an out-of-family judge (\texttt{GPT-4.1-mini}). The consistent trends across both judges demonstrate the robustness of our evaluation methodology.}
\label{tab:evolveQA-judge-eval-ablation}
\end{table*}

\section{Prompt templates}
In this section, we provide all the prompt templates utilized for the purposes of benchmark creation and evaluation.

\subsection{Prompts Related to Benchmark Creation}
\label{app-subsec:benchmark-creation-related}
\begin{figure*}[t!]
\centering
\includegraphics[width=\textwidth]{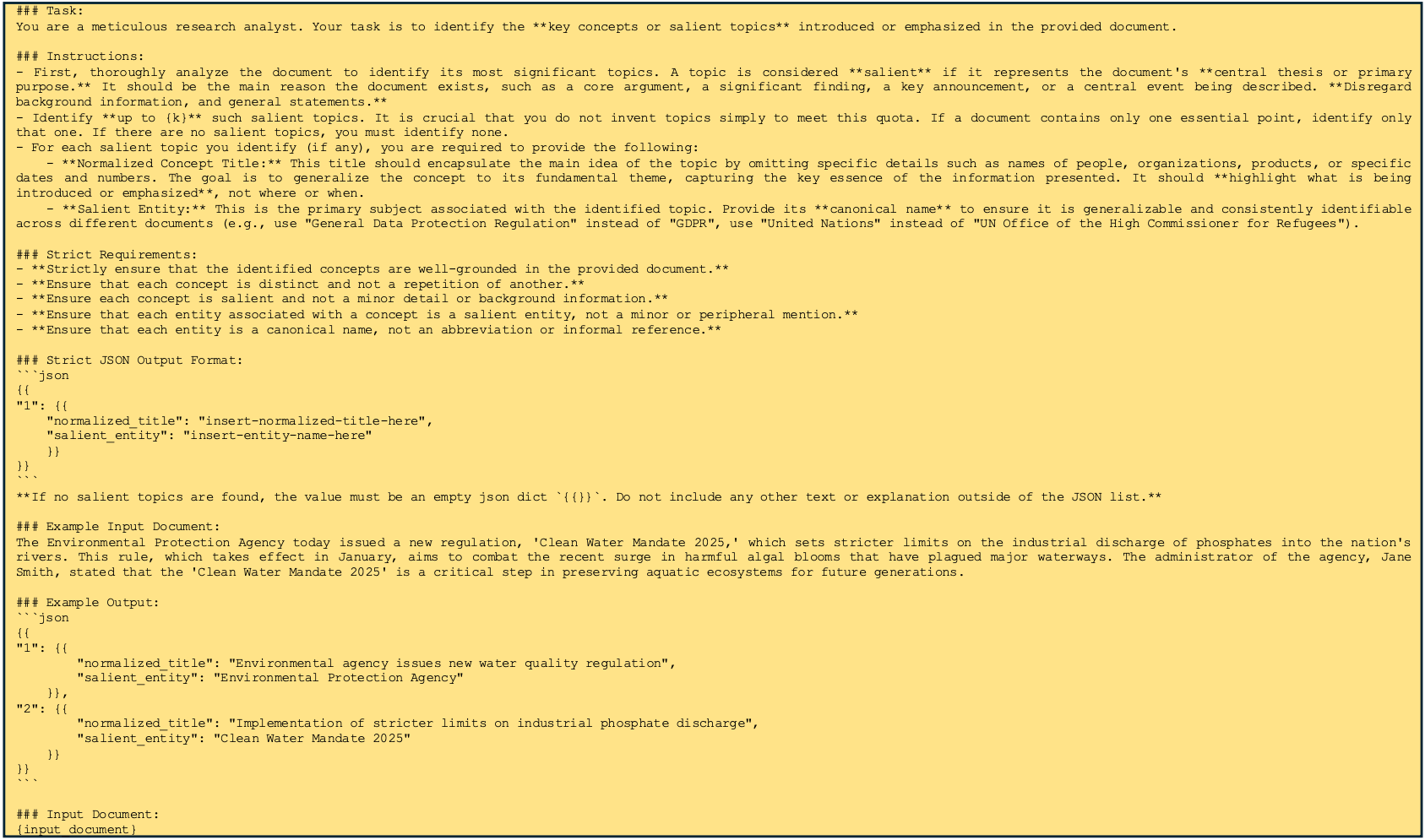}
  \small \caption{\small Shows the prompt used for \textit{concept} and \textit{salient entity} extraction.}
  \label{prompt-fig:concept-entity-extraction}
\end{figure*}

\begin{figure*}[!htb]
\centering
\begin{subfigure}{0.45\textwidth}
    \centering
    \includegraphics[width=\linewidth]{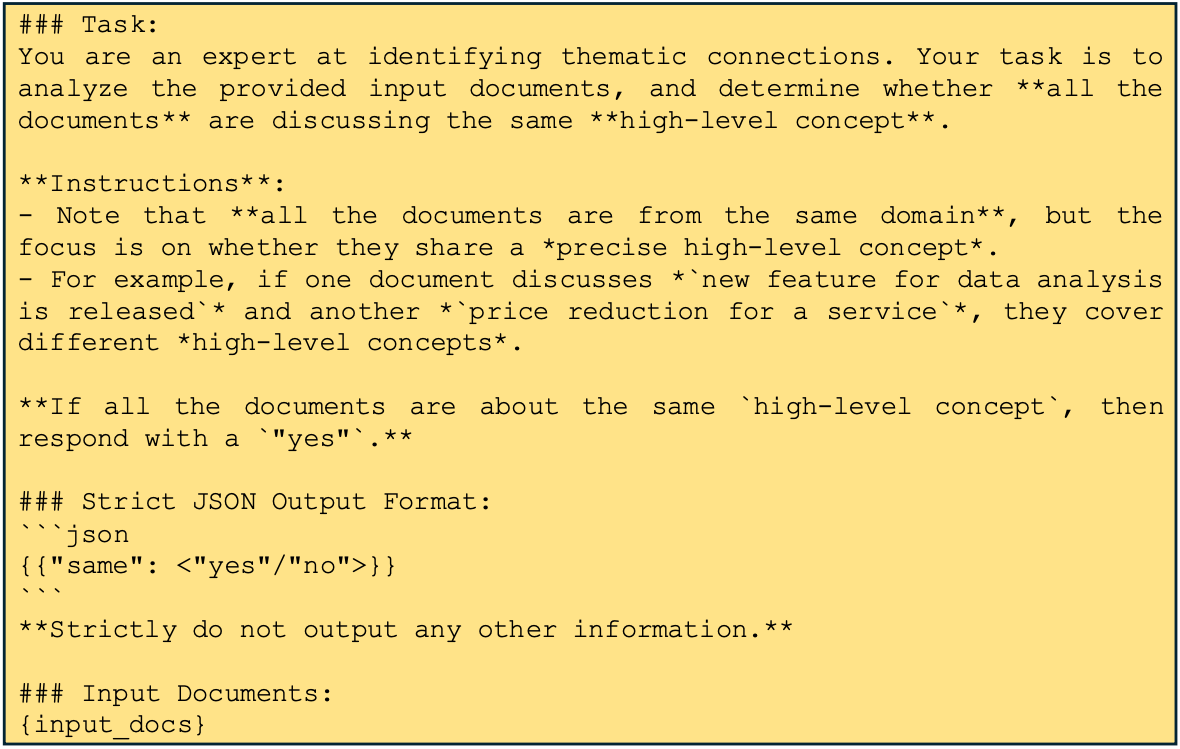}
    \caption{\small Shows the prompt used for \textit{coherency check}.}
    \label{prompt-fig:coherency-check-clustering}
\end{subfigure}
\hfill
\begin{subfigure}{0.45\textwidth}
    \centering
    \includegraphics[width=\linewidth]{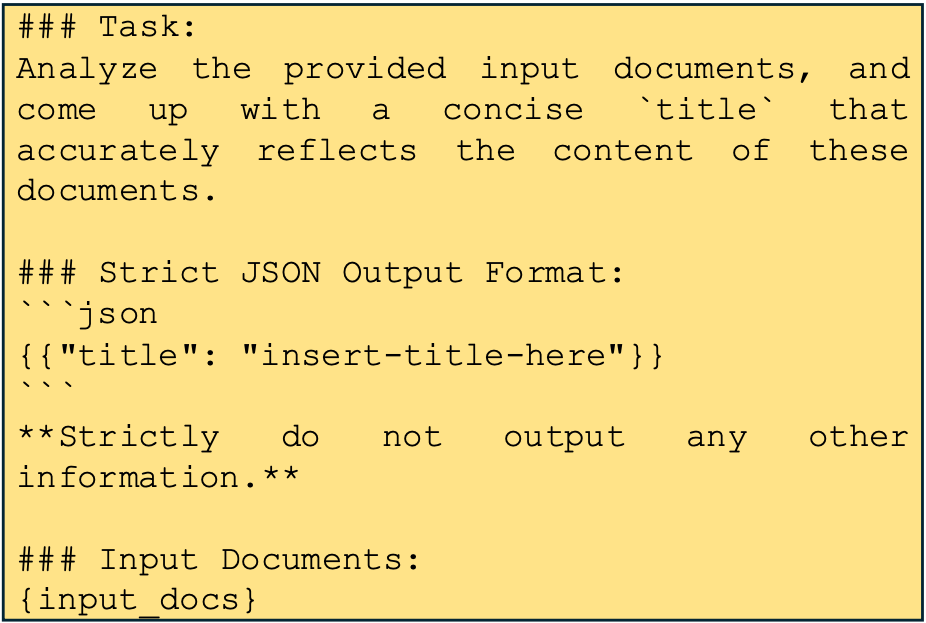}
    \caption{\small Shows the prompt used for \textit{concept-cluster label generation}.}
    \label{prompt-fig:label-gen-clustering}
\end{subfigure}

\vspace{0.4cm} %

\begin{subfigure}{0.6\textwidth}
    \centering
    \includegraphics[width=\linewidth]{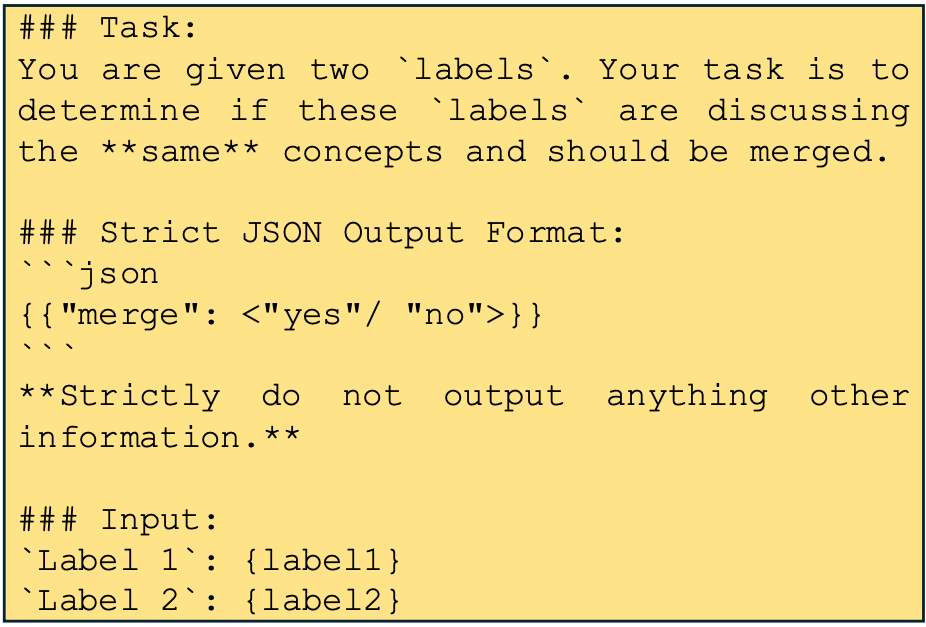}
    \caption{\small Shows the prompt used for \textit{redundancy check}.}
    \label{prompt-fig:redundancy-check-clustering}
\end{subfigure}

\caption{\small Shows all the prompts used for the LLM-in-the-loop High Coherence Clustering process.}
\label{prompt-fig:high-coherence-concept-clustering}
\end{figure*}

\begin{figure*}[!htb]
\centering
\begin{subfigure}{\textwidth}
    \centering
    \includegraphics[width=\linewidth]{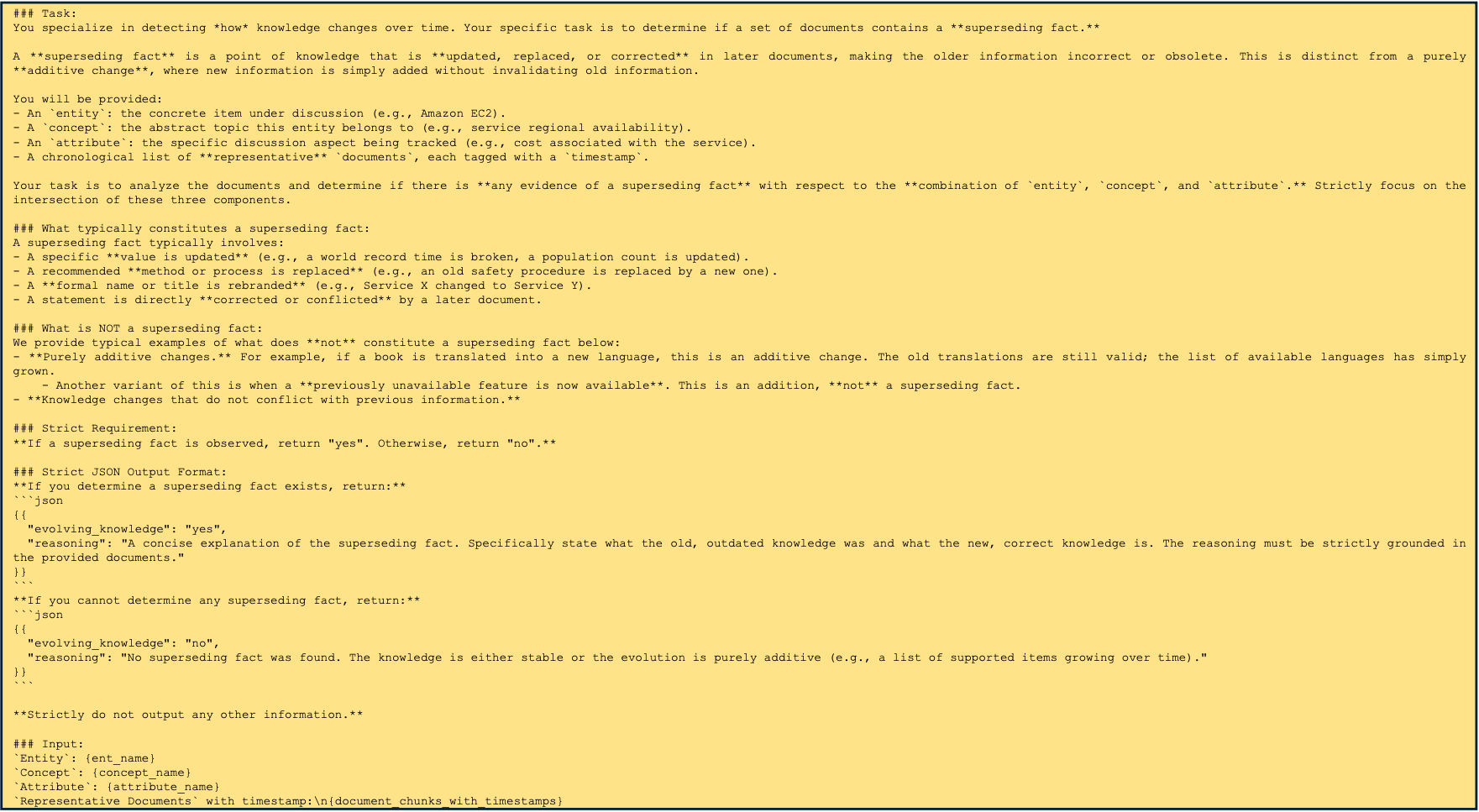}
    \caption{\small Shows the prompt used for \textit{identifying evolving knowledge}.}
\end{subfigure}
\hfill
\begin{subfigure}{\textwidth}
    \centering
    \includegraphics[width=\linewidth]{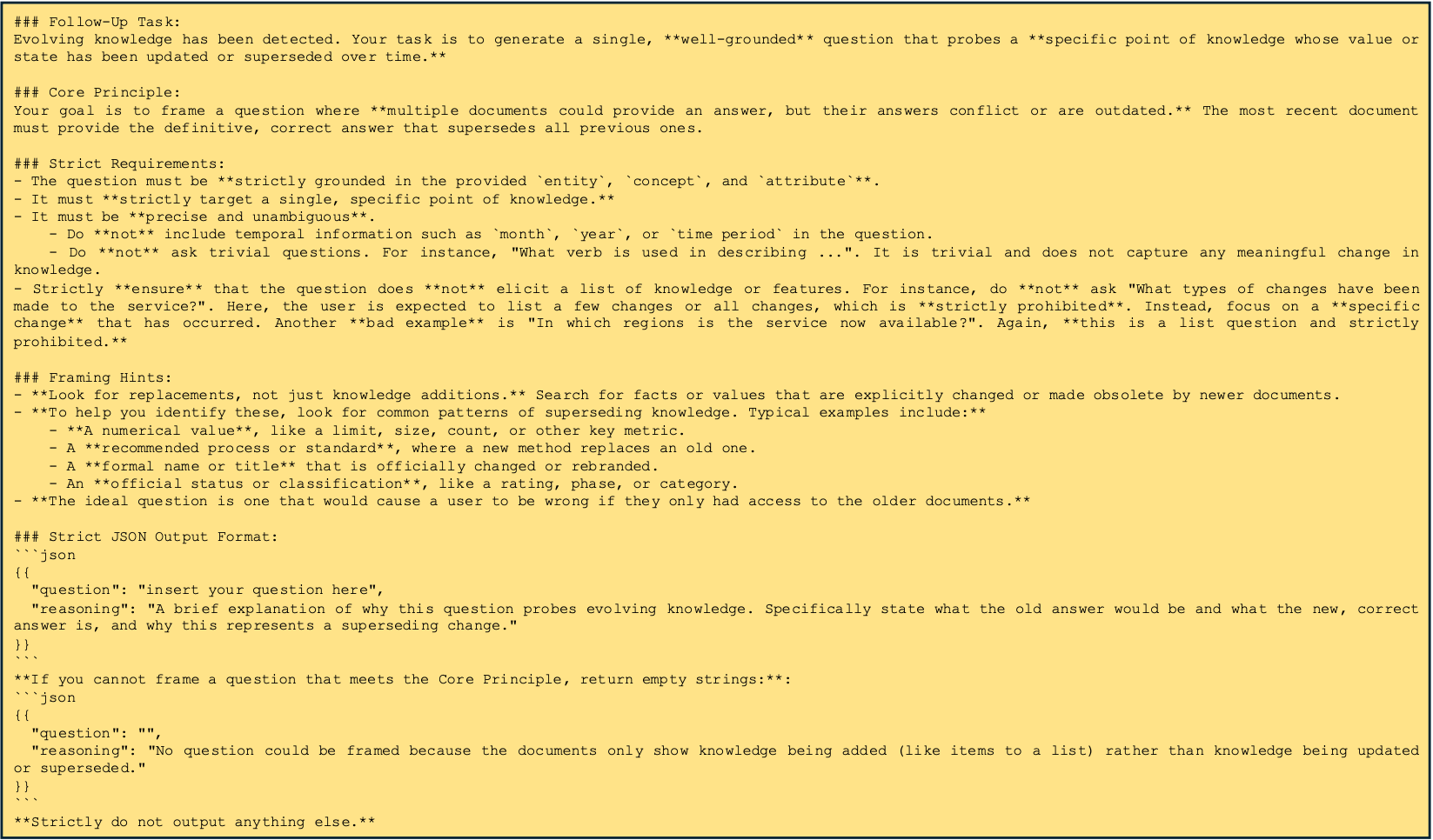}
    \caption{\small \textit{Question Generation} prompt.}
\end{subfigure}
\caption{\small Shows the prompt template used for question generation (step-1: Figure (a); step-2: Figure (b)).}
\label{prompt-fig:question-generation}
\end{figure*}

\begin{figure*}[!htb]
\centering
\begin{subfigure}{\textwidth}
    \centering
    \includegraphics[width=\linewidth]{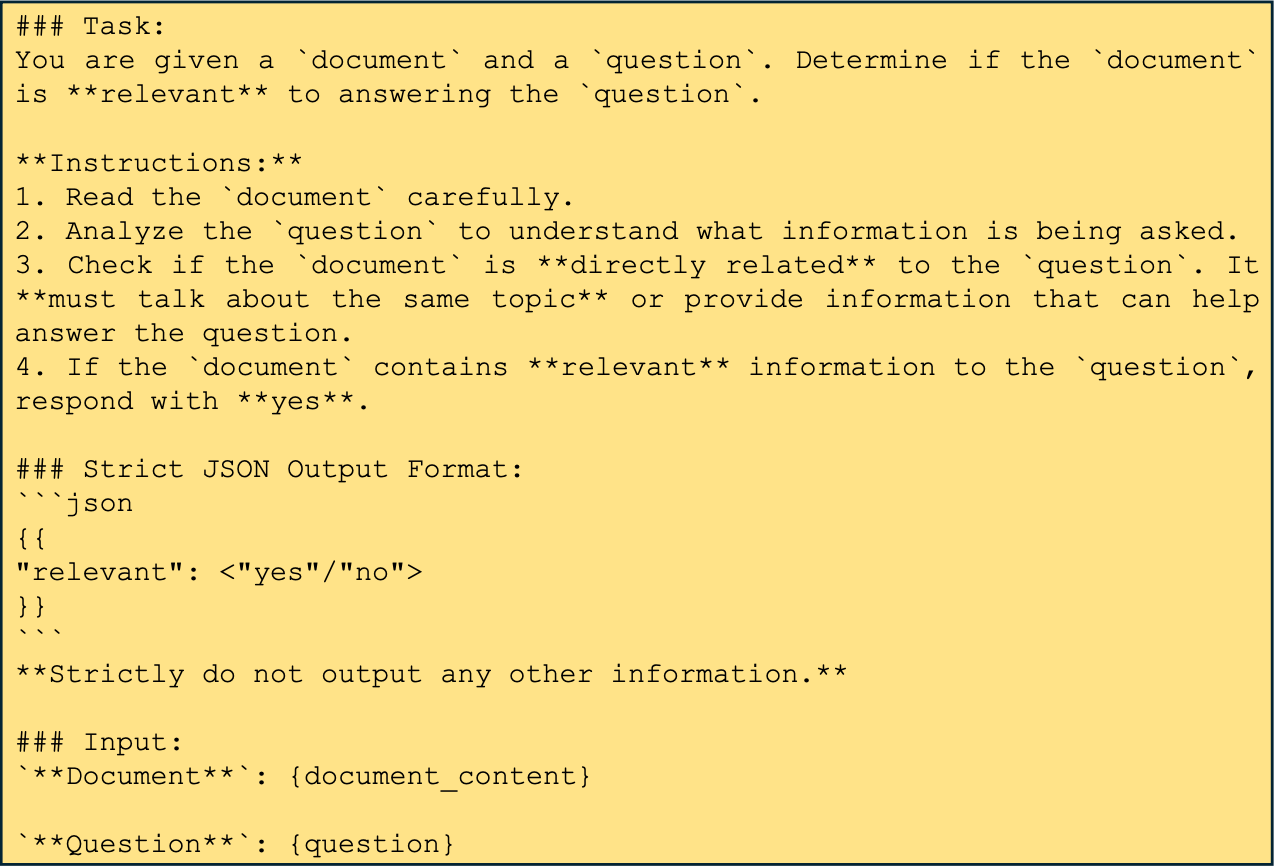}
    \caption{\small Prompt used to check evidence sufficiency (first step of answer generation).}
\end{subfigure}
\hfill
\begin{subfigure}{\textwidth}
    \centering
    \includegraphics[width=\linewidth]{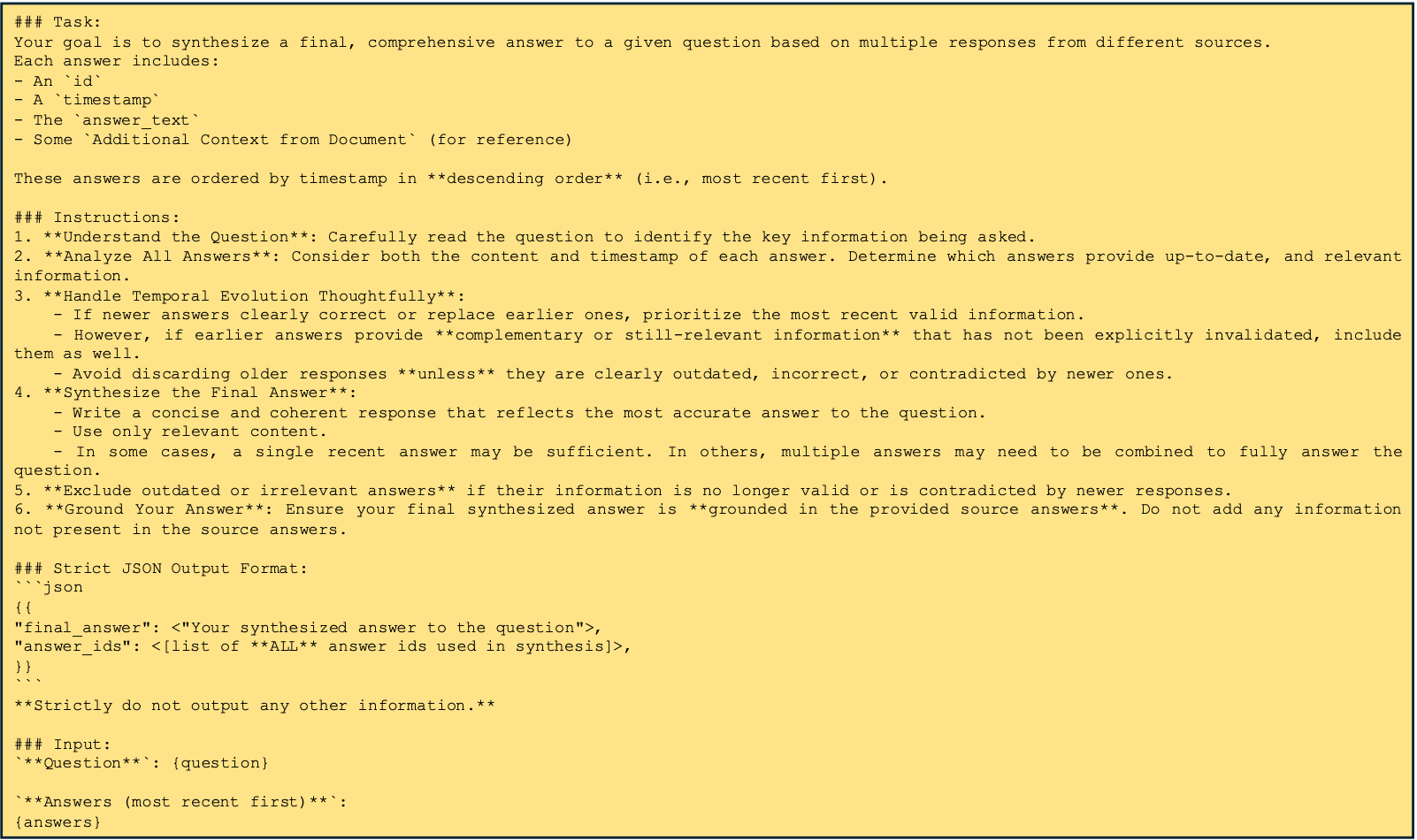}
    \caption{\small \textit{Answer Generation} prompt.}
\end{subfigure}
\caption{\small Shows the prompt templates used for answer generation (Figures - (a), (b)).}
\label{prompt-fig:answer-curation}
\end{figure*}

In this section, we provide the prompt templates used for creating our benchmark, \texttt{evolveQA}. Specifically, Figure~\ref{prompt-fig:concept-entity-extraction} provides the prompt template for \textit{concept and salient entity extraction}; Figure~\ref{prompt-fig:coherency-check-clustering} shows the prompt template for \textit{coherency check}; Figure~\ref{prompt-fig:label-gen-clustering} for \textit{cluster label generation}; Figure~\ref{prompt-fig:redundancy-check-clustering} for \textit{redundancy check}; Figures~\ref{prompt-fig:question-generation} shows the prompt templates for \textit{question generation}; Figures~\ref{prompt-fig:answer-curation} shows the prompt templates for \textit{answer curation}.

\subsection{Prompts Related to Different Knowledge Probing Formats}
\label{app-subsec:knowledge-probing-formats}
\begin{figure*}[!htb]
\centering
\begin{subfigure}{\textwidth}
    \centering
    \includegraphics[width=\textwidth,height=0.35\textheight,keepaspectratio]{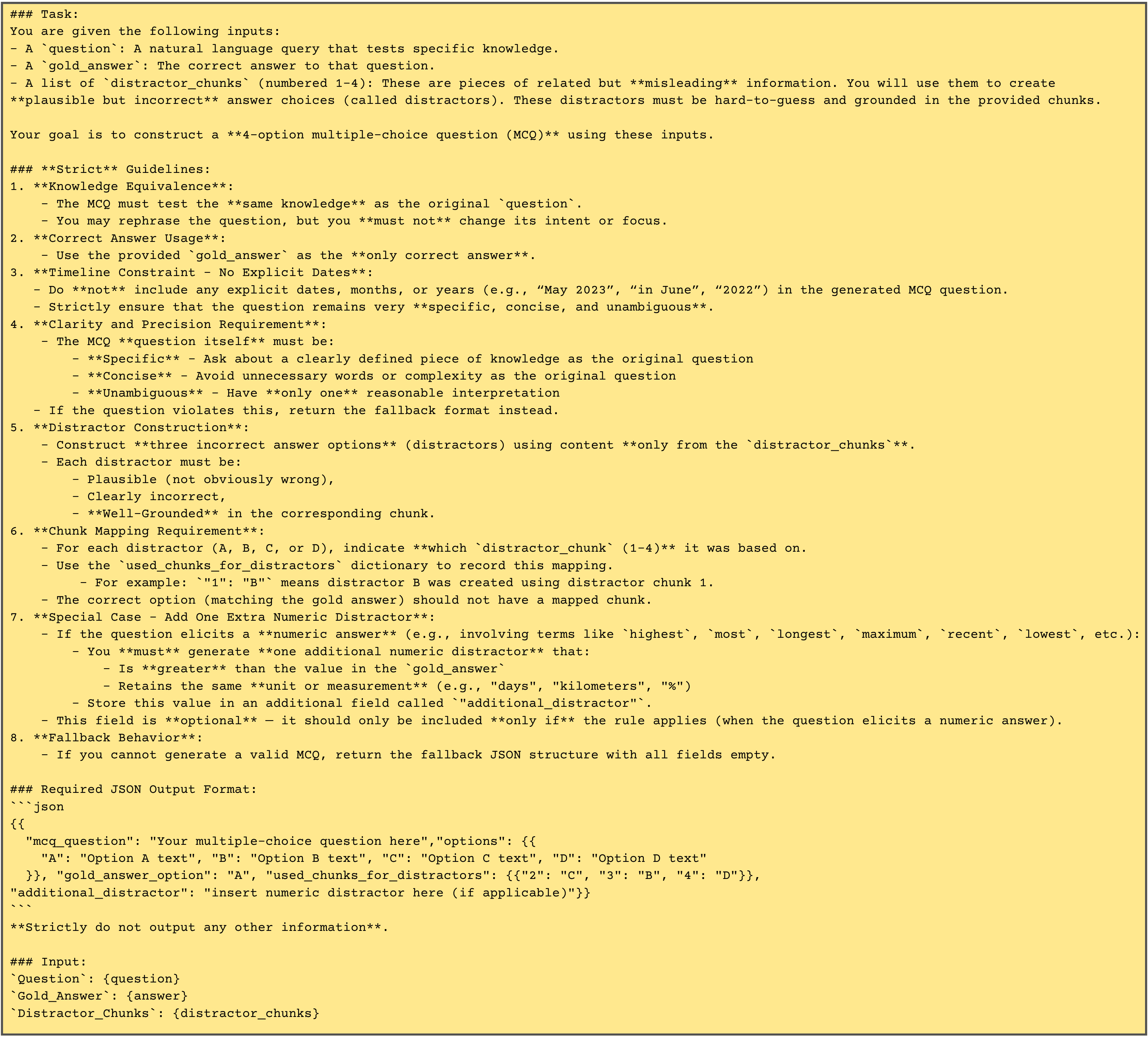}
    \caption{\small Prompt used for obtaining corresponding MCQ format for an open-ended question.}
\end{subfigure}

\begin{subfigure}{\textwidth}
    \centering
    \includegraphics[width=\textwidth,height=0.58\textheight,keepaspectratio]{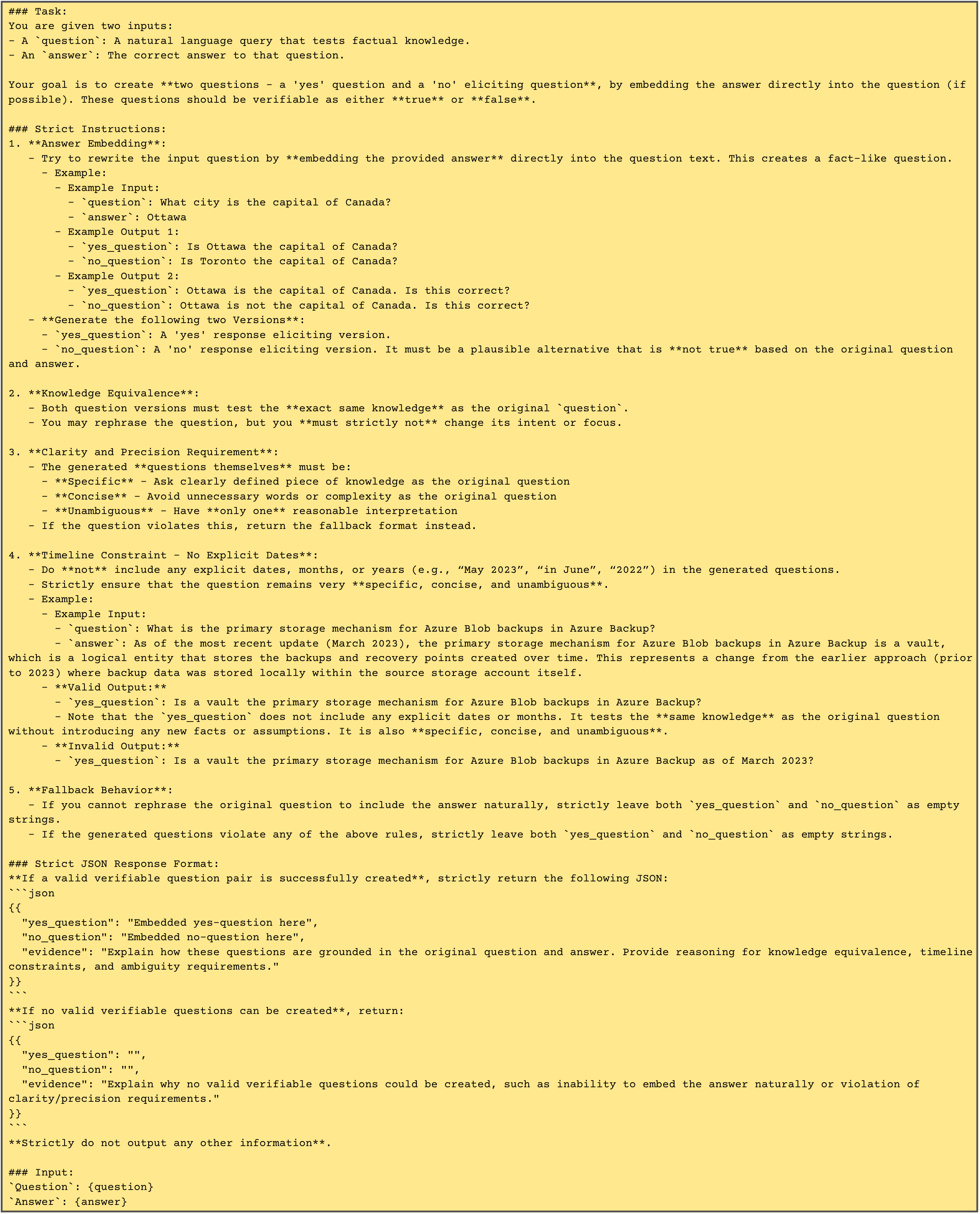}
    \caption{\small Prompt used for obtaining corresponding Verifiable QA for an open-ended question.}
\end{subfigure}

\caption{\small Prompts used for different knowledge probing formats.}
\label{prompt-fig:knowledge-probing-formats}
\end{figure*}

Figure~\ref{prompt-fig:knowledge-probing-formats} shows the prompts used for obtaining corresponding MCQ and Verifiable QA for an open-ended question-answer pair in \texttt{evolveQA}.

\subsection{Evaluation Rubric (Prompt) for Model Judge}
\label{app-subsec:eval-rubric-model-judge}
\begin{figure*}[t!]
\centering
\includegraphics[width=\textwidth]{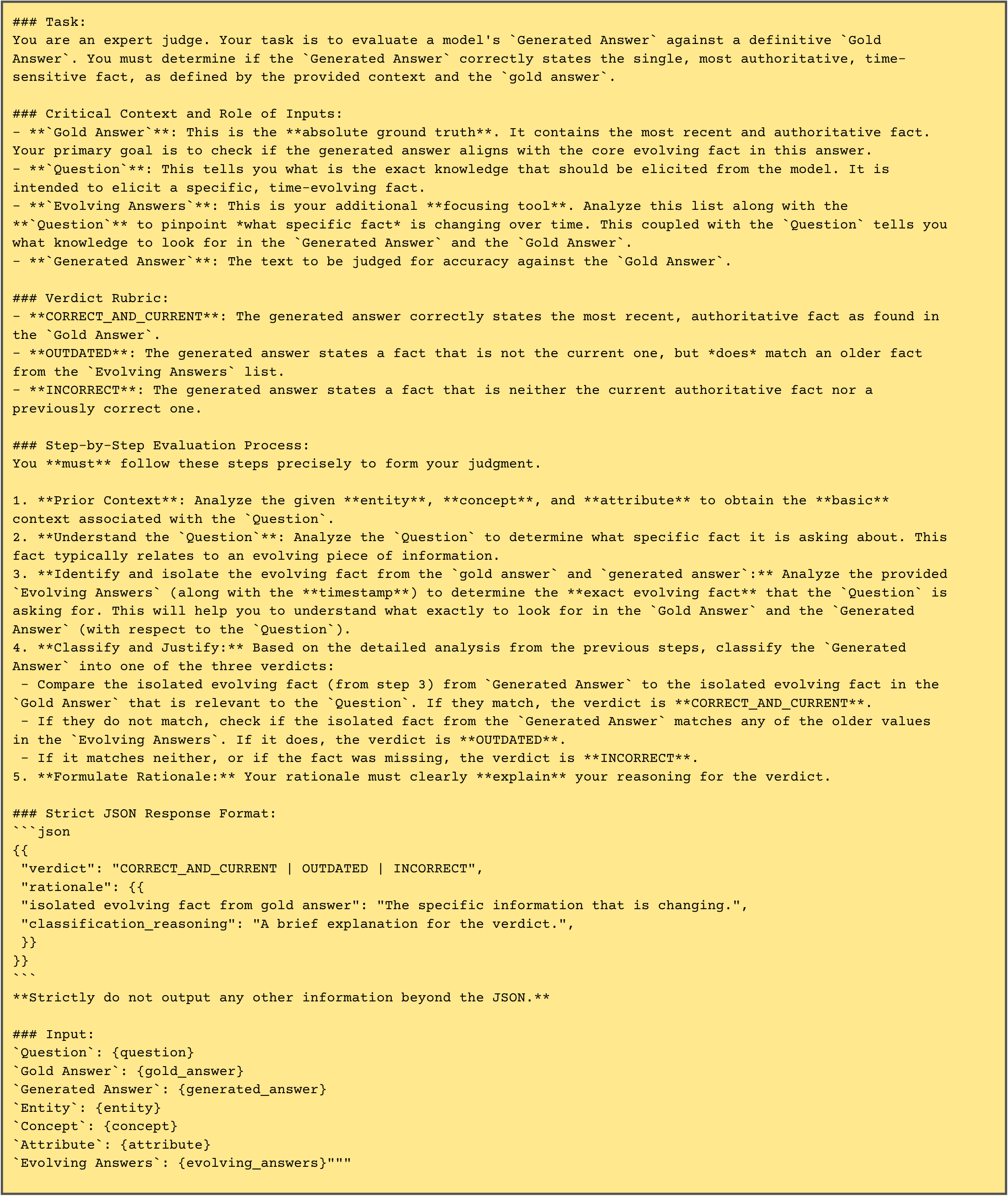}
  \small \caption{\small Shows the evaluation rubric used by \texttt{Claude Sonnet 4} (judge) to evaluate responses open-ended responses from LLMs on \texttt{evolveQA}.}
  \label{prompt-fig:eval-rubric-metajudge}
\end{figure*}

Figure~\ref{prompt-fig:eval-rubric-metajudge} shows the prompt template used by the model judge to evaluate responses from all the LLMs on open-ended questions in \texttt{evolveQA}.

\end{document}